\documentclass[times,twocolumn,final]{elsarticle}
\usepackage{epsfig}
\usepackage{graphicx}
\usepackage{amsmath}
\usepackage{amssymb}
\usepackage{comment}
\usepackage{algorithm}
\usepackage{algpseudocode}
\usepackage{enumitem}
\usepackage{float}
\usepackage{lineno}
\usepackage{subfig}
\usepackage{tabularx}
\usepackage{multirow}
\usepackage{xspace}

\usepackage{yjvci}
\usepackage{framed,multirow}

\usepackage{url}
\usepackage{xcolor}
\usepackage{hyperref}

\def\modelname{IPVNet} 
\newcolumntype{Y}{>{\centering\arraybackslash}X}

\definecolor{newcolor}{rgb}{.8,.349,.1}

\journal{Journal of Visual Communication and Image Representation}

\begin{document}

\begin{frontmatter}

\title{IPVNet: Learning Implicit Point-Voxel Features for Open-Surface 3D
Reconstruction
}%

\author[1]{Mohammad Samiul \snm{Arshad}}
\author[1]{William J. \snm{Beksi}\corref{cor1}}
\cortext[cor1]{Corresponding author: william.beksi@uta.edu (W.J. Beksi). 
}

\address[1]{The University of Texas at Arlington, 
701 S. Nedderman Drive,
Arlington, TX, 76019, 
USA}


\begin{abstract}
Reconstruction of 3D open surfaces (e.g., non-watertight meshes) is an
underexplored area of computer vision. Recent learning-based implicit techniques
have removed previous barriers by enabling reconstruction in arbitrary
resolutions. Yet, such approaches often rely on distinguishing between the
\textit{inside} and \textit{outside} of a surface in order to extract a zero
level set when reconstructing the target. In the case of open surfaces, this
distinction often leads to artifacts such as the artificial closing of surface
gaps. However, real-world data may contain intricate details defined by salient
surface gaps. Implicit functions that regress an unsigned distance field have
shown promise in reconstructing such open surfaces.  Nonetheless, current
unsigned implicit methods rely on a \textit{discretized representation} of the
raw data. This not only bounds the learning process to the representation's
resolution, but it also introduces outliers in the reconstruction. To enable
accurate reconstruction of open surfaces without introducing outliers, we
propose a learning-based implicit point-voxel model (\modelname). \modelname\
predicts the unsigned distance between a surface and a query point in 3D space
by leveraging both raw point cloud data and its discretized voxel counterpart.
Experiments on synthetic and real-world public datasets demonstrates that
\modelname\ \textit{outperforms} the state of the art while producing
\textit{far fewer} outliers in the resulting reconstruction. 
\end{abstract}


\end{frontmatter}


\section{Introduction}
\label{sec:introduction}
\begin{figure*}
\centering
\subfloat[]{\includegraphics[width=0.35\linewidth,height=0.22\linewidth]{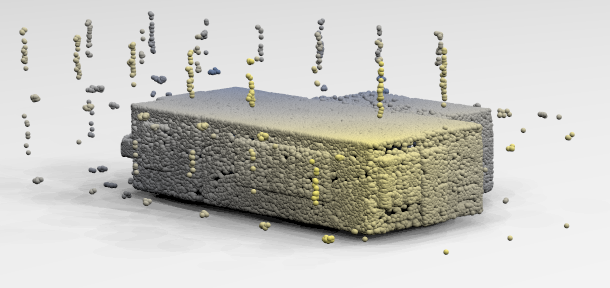}}
\subfloat[]{\includegraphics[width=0.35\linewidth,height=0.22\linewidth]{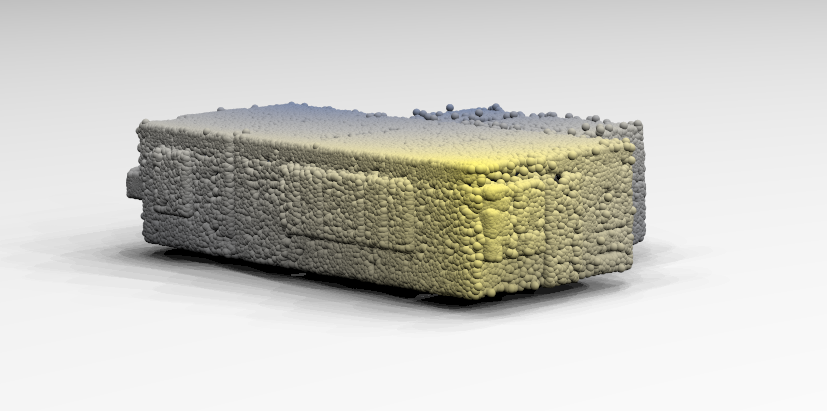}}
\caption{An outside view of a dense 3D reconstructed scene from the Gibson
Environment \cite{xia2018gibson} dataset using (a) the state of the art
\cite{chibane2020neural}, and (b) our proposed approach. Note that our method
produces significantly less outliers.}
\label{fig:short_result}
\end{figure*}
3D computer vision for generating (e.g., \cite{arshad2020progressive,
son2021progressive}) and reconstructing (e.g., \cite{pepe2022data,
kim2023pointinverter}) point clouds has gained momentum due to applications
such as robotics, autonomous driving, and virtual reality. Capturing detailed
point cloud data from the real world is a difficult and expensive task.
Moreover, due to the limitations of 3D sensor technologies (e.g., LiDAR, RGBD,
etc.), data can be sparse (i.e., missing details) and incomplete (i.e., noisy
with holes and outliers). The 3D reconstruction of missing parts and
reintroduction of details is not a trivial task. Researchers have looked into a
myriad of ways to complete 3D data. Learning-based implicit functions have
become popular among 3D reconstruction techniques due to their demonstrated
superiority in capturing details and the ability to generate data in arbitrary
resolutions.

Implicit functions operate by first converting raw data into an occupancy grid
and then learning a voxel occupancy or a distance field that classifies a query
point as either inside or outside of the surface. In low resolutions, occupancy
grids lose information during voxelization since multiple points within the
boundary of a grid are merged together. To preserve fine details in the input
data, a high-resolution representation is required. However, the computational
costs and memory requirements increase \textit{cubically} with voxel resolution.
For example, Chibane et al. \cite{chibane2020neural} require 8.86 GB of memory
to train with a single input (batch size 1) at a resolution of $256^3$. This
large memory footprint makes it impractical to scale beyond the aforesaid
resolution.

Instead of relying on the voxels, researchers have also tried to use raw point
clouds with a learned signed distance field (SDF) on the surface. Nevertheless,
implicit functions that learn an SDF via extraction of a zero level set must
distinguish between the inside/outside of the surface. As a result, the
reconstruction is produced as a closed surface even if the target shape includes
surface gaps. However, real-world data may consist of salient open surfaces.
Closing the surface of such data often leads to the introduction of outliers and
lost details. 

To reconstruct accurate geometry without introducing outliers, we propose
\modelname, an implicit model that learns an unsigned distance field (UDF) by
jointly accumulating features from raw point clouds and voxel grids to
reconstruct open surfaces. As shown in Fig.~\ref{fig:short_result}, our approach
produces significantly less outliers compared to the state of art
\cite{chibane2020neural}. Note that by reconstructing a surface, we refer to the
construction of dense point clouds that lie on the surface, which is a key part
of the reconstruction process. Although one could extract and render the surface
mesh from such point clouds, we present our results in the form of \textit{raw
point clouds}. For completeness, we record additional results as rendered meshes
in our experimental evaluation.

The technique of learning features from both point clouds and voxels has been
shown to achieve superior performance in classification and segmentation,
detection, and generation. However, to the best of our knowledge, our work is
the first approach to combine point-voxel features to learn a UDF for
open-surface reconstruction. Such improved features allow our model to
reconstruct \textit{richer detail} even with \textit{low-resolution} voxel
grids. Moreover, the involvement of raw point features allows us to use a more
sophisticated inference module that produces \textit{significantly fewer}
outliers in the reconstructed output. Our key contributions are summarized as
follows.
\begin{itemize}
  \item We introduce \modelname, a novel approach for implicitly learning from raw
  point cloud and voxel features to 3D reconstruct complex open surfaces.
  \item We develop an inference module that extracts a zero level set from a UDF
  and drastically lowers the amount of outliers in the reconstruction.
  \item We show that \modelname\ outperforms the state of the art on both
  synthetic and real-world public datasets, and we provide an ablation analysis to
  understand the importance of point-voxel fusion.
\end{itemize}
To reproduce and improve upon our results, the project source code is publicly
available to the research community \cite{ipvnet}.

The remainder of this paper is organized as follows. We provide an overview of
related research in Sec.~\ref{sec:related_work}. In Sec.~\ref{sec:background},
a concise summary of implicit functions is provided. The details of \modelname\
are presented in Sec.~\ref{sec:implicit_learning_with_point-voxel_features},
and the experimental evaluation and results are described in
Sec.~\ref{sec:experiments}. We discuss limitations and future directions of our
work in Sec.~\ref{sec:limitations_and_future_directions} and a conclusion is
given in Sec.~\ref{sec:conclusion}.

\section{Related work}
\label{sec:related_work}
3D reconstruction is a well researched area with a number of different
approaches and algorithms. In this section, we review and compare our work with
learning-based implicit approaches. For a more comprehensive review, we refer
the reader to a contemporary survey on 3D reconstruction \cite{you2020survey}.

\subsection{Implicit function learning}
\label{subsec:implicit_function_learning}
Instead of explicitly predicting a surface, implicit feature learning methods
try to either predict if a particular point in 3D space is inside or outside of
a target surface (occupancy), or determine how far the point is from the target
surface (distance). To reconstruct 3D data in arbitrary resolutions and learn a
continuous 3D mapping, Mescheder et al. \cite{mescheder2019occupancy} presented
a network that predicts voxel occupancy. Peng et al.
\cite{peng2020convolutional} improved the occupancy network by incorporating 2D
and 3D convolutions. An encoder-decoder architecture was used by Chen et al.
\cite{chen2019learning} to learn voxel occupancy. Michalkiewicz et al.
\cite{michalkiewicz2019deep} estimated an oriented level set to extract a 3D
surface. Littwin and Wolf \cite{littwin2019deep} used encoded feature vectors as
the network weights to predict voxel occupancy. Park et al.
\cite{park2019deepsdf} introduced DeepSDF, an encoder-decoder architecture that
predicts a signed distance to the surface instead of voxel occupancy. Genova et
al. \cite{genova2020local} divided an object's surface into a set of shape
elements and utilized an encoder-decoder to learn occupancy.  Sitzmann et al.
\cite{sitzmann2020implicit} introduced SIREN to implicitly learn complex signals
for various downstream tasks including 3D shape representations via a signed
distance function.

\begin{figure*}[ht!]
\centering
\includegraphics[scale=0.4]{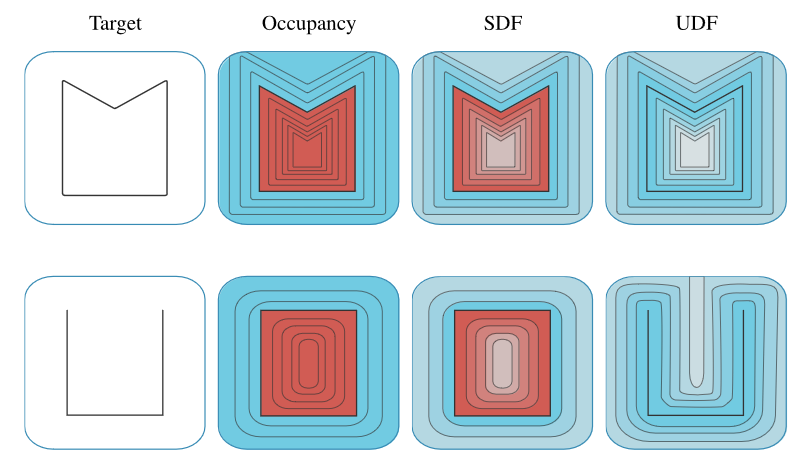}
\caption{A 2D representation of a closed (top row) and an open (bottom row)
surface reconstruction via occupancy, signed distance field (SDF), and unsigned
distance field (UDF). Note that the occupancy and SDF reconstructions of the
open surface closes the gap by producing an artifact while a UDF can preserve
the surface opening. The color intensity represents the distance from the
surface where blue represents a positive value, and red represents a zero
(occupancy) or negative value (SDF).}
\label{fig:oc_sdf_udf}
\end{figure*}

Bhatnagar et al. \cite{bhatnagar2020combining} combined implicit functions with
parametric modeling to jointly reconstruct body shape under clothing by
predicting occupancy. To retain richer details in the reconstruction, Chibane et
al. \cite{chibane2020implicit} used 3D feature tensors to predict voxel
occupancy. Rather than transforming point clouds into a occupancy grid, Atzmon
and Lipman made use of raw point clouds to learn and predict an SDF to the
target surface \cite{atzmon2020sal}, and later incorporated derivatives in a
regression loss to further improve the reconstruction accuracy
\cite{atzmon2020sald}. Gropp et al. \cite{gropp2020implicit} used geometric
regularization to learn directly from raw point clouds. To make the
reconstruction process more scalable, Mi et al. \cite{mi2020ssrnet} introduced
SSRNet to construct local geometry-aware features for octree vertices. By
leveraging gradient-based meta-learning algorithms, Sitzmann et al.
\cite{sitzmann2020metasdf} developed MetaSDF to improve the generalization
ability of implicit learning. 

With a similar aim of improving generalization, Tretschk et al.
\cite{tretschk2020patchnets} created a patch-based representation to learn an
SDF. A local implicit grid to learn an SDF and reconstruct 3D data was used by
Jiang et al. \cite{jiang2020local}. Liu et al. \cite{liu2021deep} implemented
deep implicit least squares to regress an SDF for 3D reconstruction.  Deep
implicit fusion to estimate an SDF for online 3D reconstruction was presented by
Huang et al. \cite{huang2021di}. Duggal et al.  \cite{duggal2021secrets} used
signed distance regression via neural implicit modeling for 3D vehicle
reconstruction from partial/noisy data. Sign agnostic learning to estimate a
signed implicit field of a local surface for 3D reconstruction was proposed by
Zhao et al. \cite{zhao2021sign}. 

All of the aforementioned works either predict a voxel occupancy or a signed
distance value for a given query point, which is inadequate to reconstruct open
surfaces. To alleviate this inadequacy, Chibane et al. \cite{chibane2020neural}
predicted a UDF from an input voxel occupancy. A similar technique to learn a
UDF for single-view garment reconstruction was used by Zhao et al.
\cite{zhao2021learning}. Venkatesh et al. \cite{venkatesh2021deep} proposed a
closest surface point representation to reconstruct both open and closed
surfaces. A new NULL sign combined with conventional in and out labels to
reconstruct a non-watertight arbitrary topology was proposed by Chen et al.
\cite{chen20223psdf}. Ye et al. \cite{ye2022gifs} leveraged the relationship
between every two points, instead of points and surfaces, to improve the
reconstruction quality of non-watertight 3D shapes. The aforementioned works
only utilize the discretized voxel representation while we make use of raw
point clouds \textit{jointly} with voxel occupancy. This enables us to
accumulate improved features and reconstruct finer details with less outliers.

\subsection{Learning from points and voxels}
Due to the convenience of using volumetric convolutions, many works have
explored voxel-based representations (e.g., \cite{xie2018learning,
le2018pointgrid,huang20193d}). However, voxel grids grow \textit{cubically} with
resolution and their memory intensiveness imposes an upper bound to the highest
resolution possible. Point clouds are memory efficient, yet it is non-trivial to
extract features from them due to their sparsity and permutation invariant
nature. Recently, researchers started combining these two representations to get
the best out of them both.

Liu et al. \cite{liu2019point} introduced PVCNN to perform classification and
segmentation by extracting features from both point clouds and voxel grids via
voxelization and de-voxelization. Fusion between voxel and point features for
3D classification was used by Li et al. \cite{li2019mvf}. Shi et al.
\cite{shi2020pv} gathered multi-scale voxel features and combined them with
point cloud keypoint features for object detection. They further improved their
results by incorporating local vector pooling \cite{shi2021pv}. Point-voxel
fusion to detect 3D objects was used by Cui et al. \cite{cui2020pvf} and Tang
et al. \cite{tang2020searching} learned a 3D model via sparse point-voxel
convolution. 

Noh et al. \cite{noh2021hvpr} accumulated point-voxel features in a single
representation for 3D object detection. PVT, a transformer-based architecture
that learns from point-voxel features for point cloud segmentation was
introduced by Zhang et al. \cite{zhang2021pvt}. Wei et al. \cite{wei2021pv} used
point-voxel correlation for scene flow estimation and Li et al.
\cite{li2021improved} utilized point-voxel convolution for 3D object detection.
Xu et al. \cite{xu2021rpvnet} introduced RPVNet for point cloud segmentation via
point-voxel fusion. Cherenkova et al. \cite{cherenkova2020pvdeconv} utilized
point-voxel deconvolution for point cloud encoding/decoding. In contrast to the
preceding works, we use a point-voxel representation to learn an implicit
function for open-surface reconstruction. To restore lost details during
voxelization, we propose a novel aggregation strategy that accumulates features
from both point clouds and voxel grids.

\section{Background}
\label{sec:background}
Implicit functions rely on one of three output choices for surface
reconstruction. Concretely, given a latent representation $z \in \mathcal{Z}$ of
a point cloud object $x \in \mathcal{X} \subset \mathbb{R}^{N \times 3}$ and a
random query point $p \in \mathbb{R}^3$, an implicit function $f$ aims to
predict the following.
\begin{enumerate}[label=(\roman*)]
  \item Occupancy, i.e., if $p$ lies inside or outside the object,
  \begin{equation}
    f(z,p): \mathcal{Z} \times \mathbb{R}^3 \rightarrow [0,1].
    \label{eq:occupancy}
  \end{equation}
  \item Signed distance, i.e., the distance from $p$ to the inside or outside
  surface of the object,
  \begin{equation}
    f(z,p): \mathcal{Z} \times \mathbb{R}^3 \rightarrow \mathbb{R}.
    \label{eq:signed_distance}
  \end{equation}
  \item Unsigned distance, i.e., the absolute distance from $p$ to any surface
  on the object,
  \begin{equation}
    f(z,p): \mathcal{Z} \times \mathbb{R}^3 \rightarrow \mathbb{R}_+.
    \label{eq:unsigned_distance}
  \end{equation}
\end{enumerate}
In \eqref{eq:occupancy} - \eqref{eq:unsigned_distance}, $\mathcal{X}$ denotes
the input space, $\mathcal{Z}$ is the latent space, and $N \in \mathbb{N}$ is
the density/resolution of the point cloud. After learning the implicit function
$f$, it may be queried multiple times to find the decision boundary (occupancy)
or the zero level set (signed and unsigned distance) thus implicitly
reconstructing the surface of the desired object. Fig.~\ref{fig:oc_sdf_udf}
provides an overview of open/closed surface reconstruction via different
implicit techniques in 2D.

\section{Implicit learning with point-voxel features}
\label{sec:implicit_learning_with_point-voxel_features}
\begin{figure*}[t]
\centering
\includegraphics[width=0.7\linewidth]{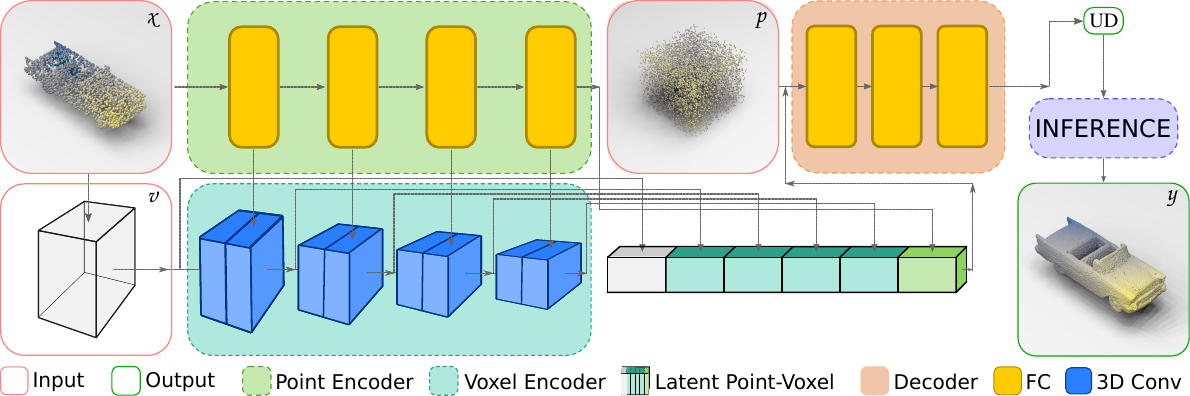}
\caption{Given a sparse point cloud $x \in \mathcal{X}$ of an object, we use a
novel encoding scheme to extract and aggregate point-voxel features from both
the raw point cloud $(x)$ and the voxel occupancy $(v)$. From the accumulated
features, a decoder module regresses the unsigned distance $UD(p,S)$ from
query point $p$ to the surface $S$. By querying the decoder multiple times,
the inference sub-module can reconstruct the surface of any target shape.}
\label{fig:model}
\end{figure*}

An overview of our network is presented in Fig.~\ref{fig:model}. We have chosen
unsigned distances as the output representation of \modelname\ due to their
precedence in representing different surfaces. Given a sparse point cloud $x \in
\mathcal{X} \subset \mathbb{R}^{N \times 3}$ of an object, we use a novel
encoding scheme to extract and aggregate point-voxel features from both the raw
point cloud $(x)$ and the voxel occupancy $(v)$. From the accumulated features,
a decoder module regresses the unsigned distance $UD(p,S)$ via any query point
$p$ to the surface $S$. In the following subsections we describe the elements of
our approach. 

\subsection{Point-voxel features}
\label{subsec:point_encoding}
To extract a set of multi-level features from a point cloud $x$, we define a
neural function
\begin{equation}
  \Theta(x) := (z_x^1, \ldots, z_x^j)\: | \: \Theta \colon \mathbb{R}^3 \rightarrow \mathcal{Z},
\end{equation}
where $z_x \in \mathcal{Z} \subset \mathbb{R}$ corresponds to the extracted
feature vector from the raw point cloud $x$, and $j$ is the total number of
layers in $\Theta$. During the early stages (i.e., when $j=1$) the encoder is
more focused on local details, whereas at the later stages the focus is shifted
towards the global structure. ReLu \cite{nair2010rectified} nonlinearity is
used for all layers except the output layer of the point encoder.

Instead of limiting the encoded features to a single dimensional vector, a
voxel representation allows for the construction of a multi-dimensional latent
matrix. However, such an encoding scheme requires the input point cloud $x$ to
be discretized into a voxel grid $v$, i.e., $x \approx v \colon \mathbb{R}^{N
\times 3} \rightarrow \mathbb{R}^{M \times M \times M}$ where $M \in
\mathbb{N}$ is the grid resolution. Due to the discretization process, voxel
grids lose information since multiple points may lie within the same voxel. To
reintroduce lost details, we combine voxel features with point features $z_x$.
Although, a different fusion strategy can be applied to combine these features,
we empirically found that a simple concatenation strategy works best.

Let $\Phi \colon \mathbb{R}^{M \times M \times M} \rightarrow \mathcal{Z}^{M
\times M \times M}$ be a neural function that encodes the combined point-voxel
features into a set of multi-dimensional feature grids $z_{xv}$ of monotonically
decreasing dimension. Then,
\begin{equation}
  \Phi(v \odot \Theta(x)) := (z_{xv}^{k \times k \times k}, \ldots, z_{xv}^{l \times l \times l}),
\end{equation}
where $k, l \in \mathbb{N}$ represents the dimensional upper and lower bound of
the feature grid $(M > k \gg l > 1)$. The subscript ${xv}$ denotes the
dependency on both points and voxels. Similar to its point counterpart, the
voxel encoder is more directed towards local details at the early stages.
However, as the dimensionality is reduced and the receptive field grows larger,
the aim shifts to the global structure. ReLu is utilized to ensure nonlinearity
and batch normalization \cite{ioffe2015batch} provides stability while training.
The latent point ($z_x^j$) from the point encoder, along with multi-dimensional
features ($z_{xv}$) from the point-voxel encoder and the discretized voxel grid
($v$), are then used to construct the latent point-voxel:
\begin{equation}
  z = \{z_x^j, \Phi(v \odot \Theta(x)), v\}.
\end{equation}

\subsection{Implicit decoding}
\label{subsec:implicit_decoding}
Given a query point $p \in \mathbb{R}^3$, a set of deep features $F_p$ is
sampled from the latent point-voxel features $z$ via a sampling function
$\Omega$ \cite{jaderberg2015spatial}. Specifically,
\begin{equation}
  \Omega(z, p) := (F_p^1 \times \dots \times F_p^{n}),
\end{equation}
where $n = \lvert z \rvert$. We extract features from a neighborhood of distance
$d \in \mathbb{R}$ along the Cartesian axes centered at $p$ to obtain rich
features. More formally,
\begin{equation}
  p := \{p + q \cdot c_i \cdot d\} \in \mathbb{R}^3 \; | \; q \in \{1,0,-1\}, i \in \{1,2,3\},
\end{equation}
where $c_i \in \mathbb{R}^3$ is the $ith$ Cartesian axis unit vector. We define
a neural function $\Psi$ that regresses the unsigned distance to the surface $S$
of $x$ from the deep features ($F_p$). Concretely,
\begin{equation}
  \Psi(F_p^1,\dots,F_p^{n}) \approxeq UD(p,S) \; | \; \Psi \colon \mathcal{Z} \rightarrow \mathbb{R}_+,
\end{equation}
where $UD(\cdot)$ is a function that returns the unsigned distance from $p$ to
the ground-truth surface $S$ for any $p \in \mathbb{R}^3$. Hence, the implicit
decoder to regress the unsigned distance at a given query point $p$ is defined
as
\begin{equation}
  f_x(z,p) := (\Omega \circ \Psi)(p) \; | \; f_x \colon \mathcal{Z} \times \mathbb{R}^3 \rightarrow \mathbb{R}_+.
\end{equation}

\subsection{Training}
\label{subsec:implicit_learning}
\modelname\ requires a pair $\{X_i, S_i\}^T_{i=1}$ associated with input $X_i$
and corresponding ground-truth surface $S_i$ for implicit learning.
Parameterized by the neural parameter $w$, the point-encoder, voxel-encoder, and
decoder are jointly trained with a mini-batch loss,
\begin{equation}
  \mathcal{L}_\mathcal{B} := \Sigma_{x \in B} \Sigma_{p \in \mathcal{P}} |\min(f_x^w(p), \delta) - \min(UD(p,S_x), \delta)|,
\end{equation}
where $\mathcal{B}$ is a mini-batch of input and $\mathcal{P} \in \mathbb{R}^3$
is a set of query points within distance $\delta$ of $S_i$. We use a clamped
distance $0 < \delta < 10$ (cm) to improve the capacity of the model to
represent the vicinity of the surface accurately.

\subsection{Surface inference}
\label{subsec:surface_inference}
We use an iterative strategy to extract surface points from $f_x$. More
specifically, given a perfect approximator $f_x(p)$ of the true unsigned
distance $UD(p,S_i)$, the projection of $p$ onto the surface $S_i$ can be
obtained by
\begin{equation}
  q := p - f_x(p) \cdot \nabla_pf_x(p), \quad q \in S_i \subset \mathbb{R}^d, \forall p \in \mathbb{R}^d/C.
  \label{eq:udf_surface_point}
\end{equation}
In \eqref{eq:udf_surface_point}, $C$ is the cut locus \cite{wolter1993cut},
i.e., a set of points that are equidistant to at least two surface points. The
negative gradient indicates the direction of the fastest decrease in distance.
In addition, we can move a distance of $f_x(p)$ to reach $q$ if the norm of the
gradient is one. By projecting a point multiple times via
\eqref{eq:udf_surface_point}, the inaccuracies due to $f_x(p)$ being an
imperfect approximator can be reduced. Furthermore, filtering the projected
points to a maximum distance threshold ($max\_thresh$) and re-projecting them
onto the surface after displacement by $d \sim \mathcal{N}(0, \delta /3)$ can
ensure higher point density within a maximum distance ($\delta$).

Instead of uniformly sampling query points within the bounding box of $S_i$, we
use the input points $X_i \in \mathbb{R}^3$ as guidance for the query points.
In particular, we apply a random uniform jitter $\mathcal{J}_a^b \in
\mathbb{R}^3$ within bounds $a$ and $b$ to displace the input points $X_i$. Due
to the inclusion of point features in learning, this procedure allows our model
to infer more accurate surface points while restricting the number of outliers.
Note that without the use of point features, this perturbation of the input
points fails to restrict the number of outliers (see
Sec.~\ref{sec:ablation_study}). The details of the inference procedure are
provided in Alg.~\ref{alg:inference}.

\begin{algorithm}
\caption{Surface Point Inference}
\label{alg:inference}
  \begin{algorithmic}[1]
  \Procedure{Inference}{$\mathcal{X}$}
    \State $\mathcal{J} \leftarrow m$ points from $\textit{U}(a, b)$
    \State $\mathcal{P}_{init} \leftarrow \{x + j\}, \forall x \in \mathcal{X}, 
    \forall j \in \mathcal{J}$
    \For{$i = 1$ to $num\_projections$}
      \State $p \leftarrow p - f_x(p) \cdot \frac{\nabla_p f_x(p)}{||\nabla_p f_x(p)||}, \forall p \in \mathcal{P}_{init}$
    \EndFor
    \State $\mathcal{P}_{filtered} \leftarrow \{p \in 
    \mathcal{P}_{init}\,|\,f_x(p) < max\_thresh\}$
    \State $\mathcal{P}_{filtered}$: draw $out\_res$ number of points with replacement
    \State $\mathcal{P}_{filtered} \leftarrow \{p + d\}\,|\,p \in 
    \mathcal{P}_{filtered}, d \sim \mathcal{N}(0, \delta /3)$
    \For{$i = 1$ to $num\_projections$}
      \State $p \leftarrow p - f_x(p) \cdot \frac{\nabla_p f_x(p)}{||\nabla_p f_x(p)||}, \forall p \in \mathcal{P}_{filtered}$
    \EndFor\\
    \hspace{5mm}\Return $\{p \in \mathcal{P}_{filtered}\,|\,f_x(p) < max\_dist\}$
  \EndProcedure
  \end{algorithmic}
\end{algorithm}

\noindent
\subsection{Implementation details}
\label{sec:implementation_details}
\modelname\ was implemented using PyTorch. To extract point cloud and voxel
features, we utilize multilayer perceptrons (MLPs) and 3D convolutional neural
networks (CNNs), respectively. Specifically, we employ a 7-layer fully-connected
MLP as the point encoder and 6-layer CNN blocks as the voxel encoder. Point
features are obtained from each of the hidden layers of the point encoder and
are combined with voxel features derived from the initial layer of each
convolution block. We use max pooling with the fully-connected layers to make
them permutation invariant.

\begin{figure*}
\centering
\includegraphics[scale=0.3]{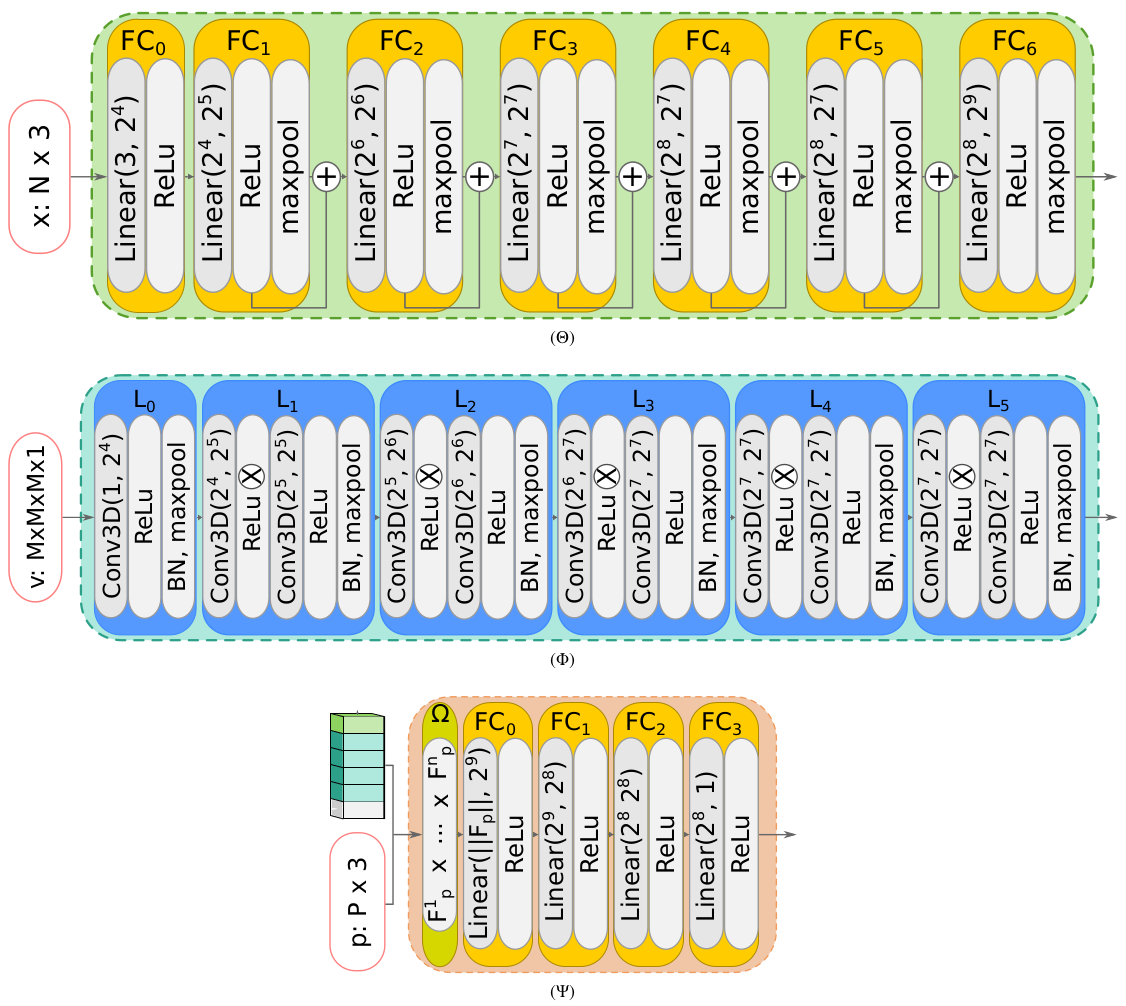}
\caption{A visual depiction of the different neural architectures of \modelname.
$\bigoplus$ in ($\Theta$) represents concatenation, and $\bigotimes$ in ($\Phi$)
indicates the fusion of point features with voxel features.}
\label{fig:network}
\end{figure*}

Fig.~\ref{fig:network} shows the details of the different neural modules of
\modelname. To train the model, we used a learning rate of $10^{-6}$ and the
Adam \cite{kingma2014adam} optimizer. With a voxel resolution of $256^{3}$ and
a batch size of 4, it takes around 3.2 seconds to perform a forward pass using
4 Nvidia GeForce GTX 1080 Ti GPUs. To infer a single target surface with a
dense point cloud consisting of 1 million points, \modelname\ takes
approximately 120 seconds.


\section{Experiments}
\label{sec:experiments}
In this section, we validate the performance of \modelname\ on the task of 3D
object and scene reconstruction from sparse point clouds. 

\subsection{Baselines and metrics}
\label{sec:baselines_and_metrics}
To compare the reconstruction quality of \modelname, we utilized the
open-source implementations of NDF \cite{chibane2020neural} and GIFS
\cite{ye2022gifs} as baseline methods. For an unbiased comparison, we trained
an NDF following the directions from \cite{chibane2020neural} on our train-test
split until a minimum validation accuracy was achieved. To quantitatively
measure the reconstruction quality, we used the \textit{chamfer-}$L_2$ distance
(CD) to measure the accuracy and completeness of the surface. The CD is defined
as
\begin{equation}
  d_{CD}(Y, Y_{gt}) = \sum_{i\in Y}^{}\min_{j \in Y_{gt}} {||i-j||}^2 + \sum_{j\in Y_{gt}}^{}\min_{i\in Y} {||j-i||}^2,
\end{equation}
where $Y_{gt} \in \mathbb{R}^{\mathcal{O} \times 3}$ is the ground-truth point
cloud, $Y \in \mathbb{R}^{\mathcal{O} \times 3}$ is the reconstructed point
cloud, and $\mathcal{O} \in \mathbb{N}$ is the point density of the ground truth
and the output. In addition, precision and recall are two metrics that have been
extensively used to evaluate 3D reconstruction results. Precision quantifies the
accuracy while recall assesses the completeness of the reconstruction. For the
ground truth $Y_{gt}$ and reconstructed point cloud $Y$, the precision of an
outcome at a threshold $d$ can be calculated as
\begin{equation*}
  P(d) = \sum_{i \in Y}^{}[\min_{j \in Y_{gt}} ||i-j|| < d].
\end{equation*}
Similarly, the recall for a given $d$ is computed as
\begin{equation*}
  R(d) = \sum_{j \in Y_{gt}}^{}[\min_{i \in Y} ||j-i|| < d].
\end{equation*}
The F-score, proposed in \cite{tatarchenko2019single} as a comprehensive
evaluation, combines precision and recall to quantify the overall reconstruction
quality. Formally, the F-score at $d$ is given by
\begin{equation*}
  F(d) = \frac{2 \cdot P(d) \cdot R(d)}{P(d) + R(d)}.
\end{equation*}
An F-score of 1 indicates perfect reconstruction.

\subsection{Object reconstruction}
\label{sec:object_reconstruction}
Due to the abundance of surface openings, we chose the ``Cars" subset of the
ShapeNet \cite{chang2015shapenet} dataset for our object reconstruction
experiment. We used a random split of $70\%$-$10\%$-$20\%$ for training,
validation, and testing, respectively. To prepare the ground truth and input
points we followed the data preparation procedure outlined in
\cite{chibane2020neural}. Additionally, we fixed the output point density to
$\mathcal{O}=1$ million to extract a smooth mesh from the point cloud using a
naive algorithm (e.g., \cite{bernardini1999ball}).

\captionsetup[subfigure]{position=top, labelformat=empty, justification=centering}
\begin{figure*}[t]
\centering
\subfloat[Input]
  {\includegraphics[width=0.16\linewidth]{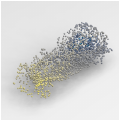}}
   \hspace{0.05pt}
\subfloat[NDF]
  {\includegraphics[width=0.16\linewidth]{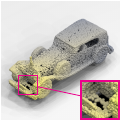}}
   \hspace{0.05pt}
\subfloat[\modelname]
  {\includegraphics[width=0.16\linewidth]{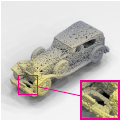}}
   \hspace{0.05pt}
\subfloat[\modelname\ (Inner View)]
  {\includegraphics[width=0.16\linewidth]{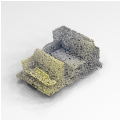}}
   \hspace{0.05pt}
\subfloat[GT]
  {\includegraphics[width=0.16\linewidth]{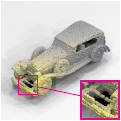}}
\par 
\subfloat
  {\includegraphics[width=0.16\linewidth]{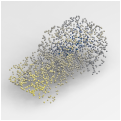}} 
  \hspace{0.05pt}
\subfloat
  {\includegraphics[width=0.16\linewidth]{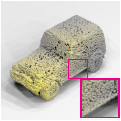}}
   \hspace{0.05pt}
\subfloat
  {\includegraphics[width=0.16\linewidth]{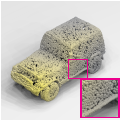}}
   \hspace{0.05pt}
\subfloat
  {\includegraphics[width=0.16\linewidth]{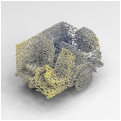}}
   \hspace{0.05pt}
\subfloat{\includegraphics[width=0.16\linewidth]{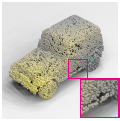}}
\par 
\subfloat
  {\includegraphics[width=0.16\linewidth]{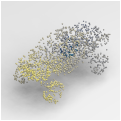}}
   \hspace{0.05pt}
\subfloat
  {\includegraphics[width=0.16\linewidth]{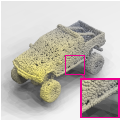}}
   \hspace{0.05pt}
\subfloat
  {\includegraphics[width=0.16\linewidth]{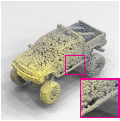}}
   \hspace{0.05pt}
\subfloat
  {\includegraphics[width=0.16\linewidth]{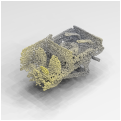}}
   \hspace{0.05pt}
\subfloat
  {\includegraphics[width=0.16\linewidth]{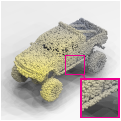}}
\caption{Object reconstruction using NDF \cite{chibane2020neural}, \modelname,
and the ground truth (GT) from the ShapeNet Cars \cite{chang2015shapenet} test
set. \modelname\ performs better on reconstructing thin structures and
preserving small gaps (inset images).}
\label{fig:result_obj_recon}
\end{figure*}

\begin{figure*}
\centering
\subfloat{
  \includegraphics[width=0.13\linewidth,height=0.2\linewidth, trim={200 0 20 0},clip]{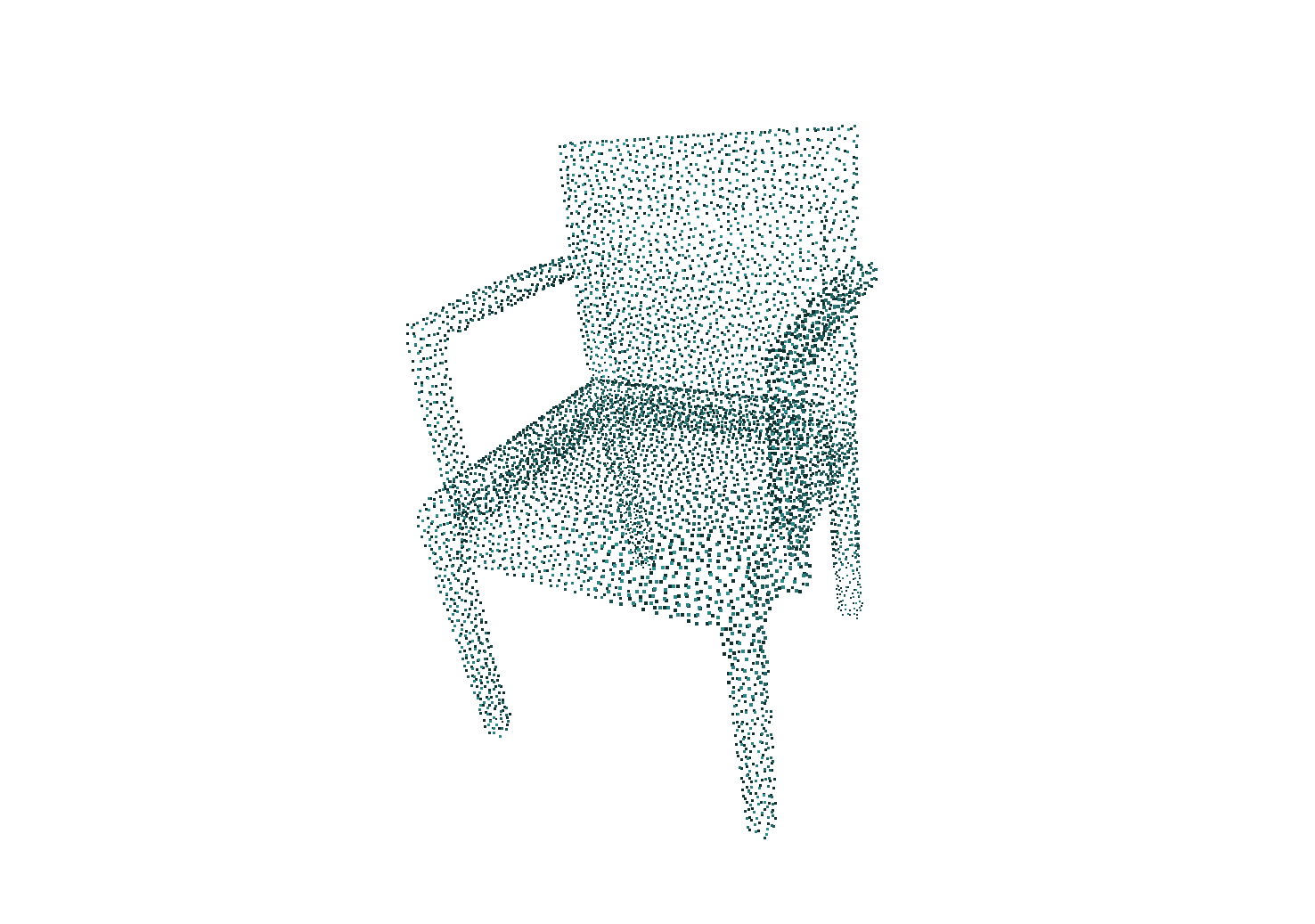} \hspace{-3em}
  \includegraphics[width=0.13\linewidth,height=0.2\linewidth, trim={200 0 20 0},clip]{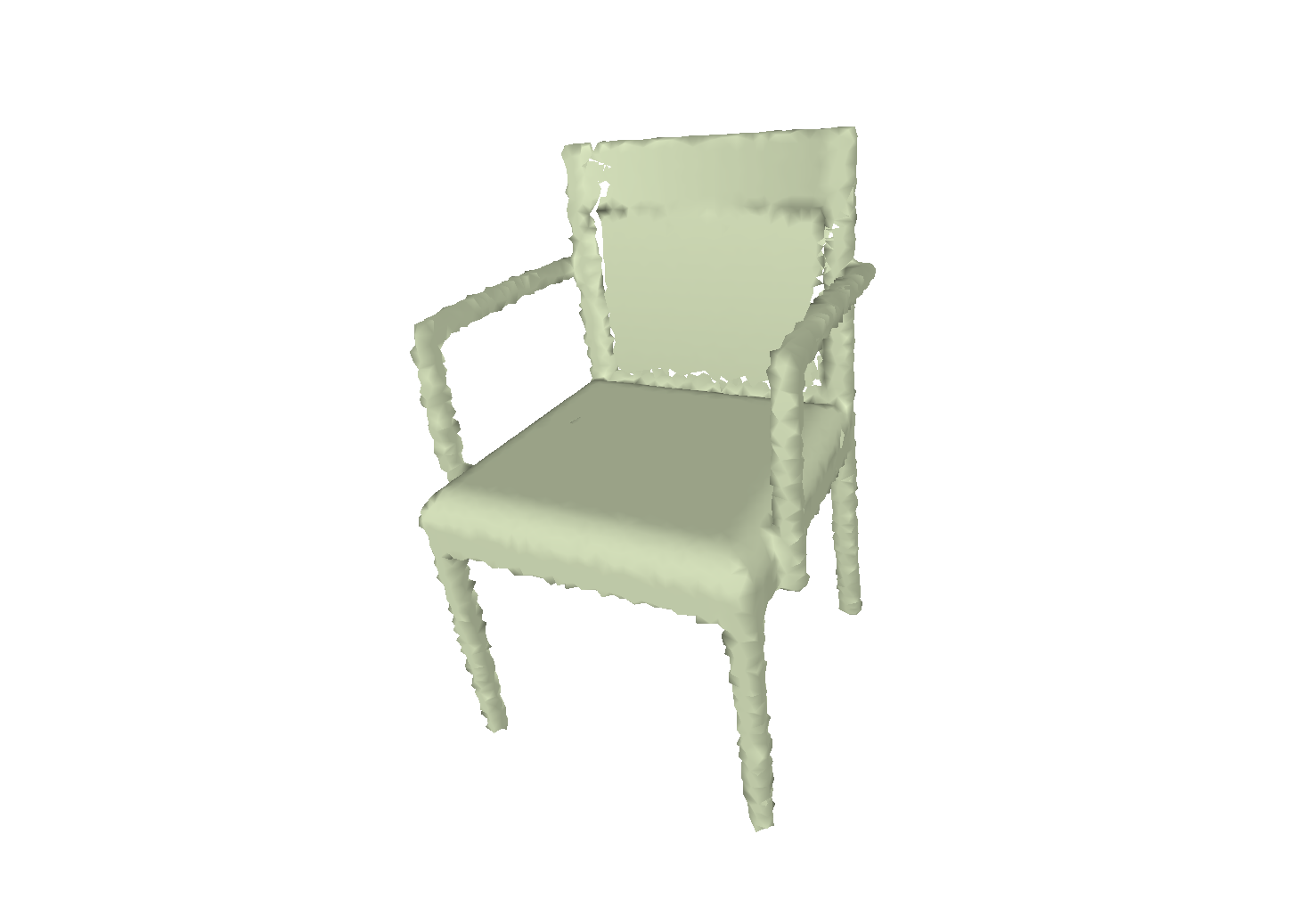} \hspace{-3em}
  \includegraphics[width=0.13\linewidth,height=0.2\linewidth, trim={200 0 20 0},clip]{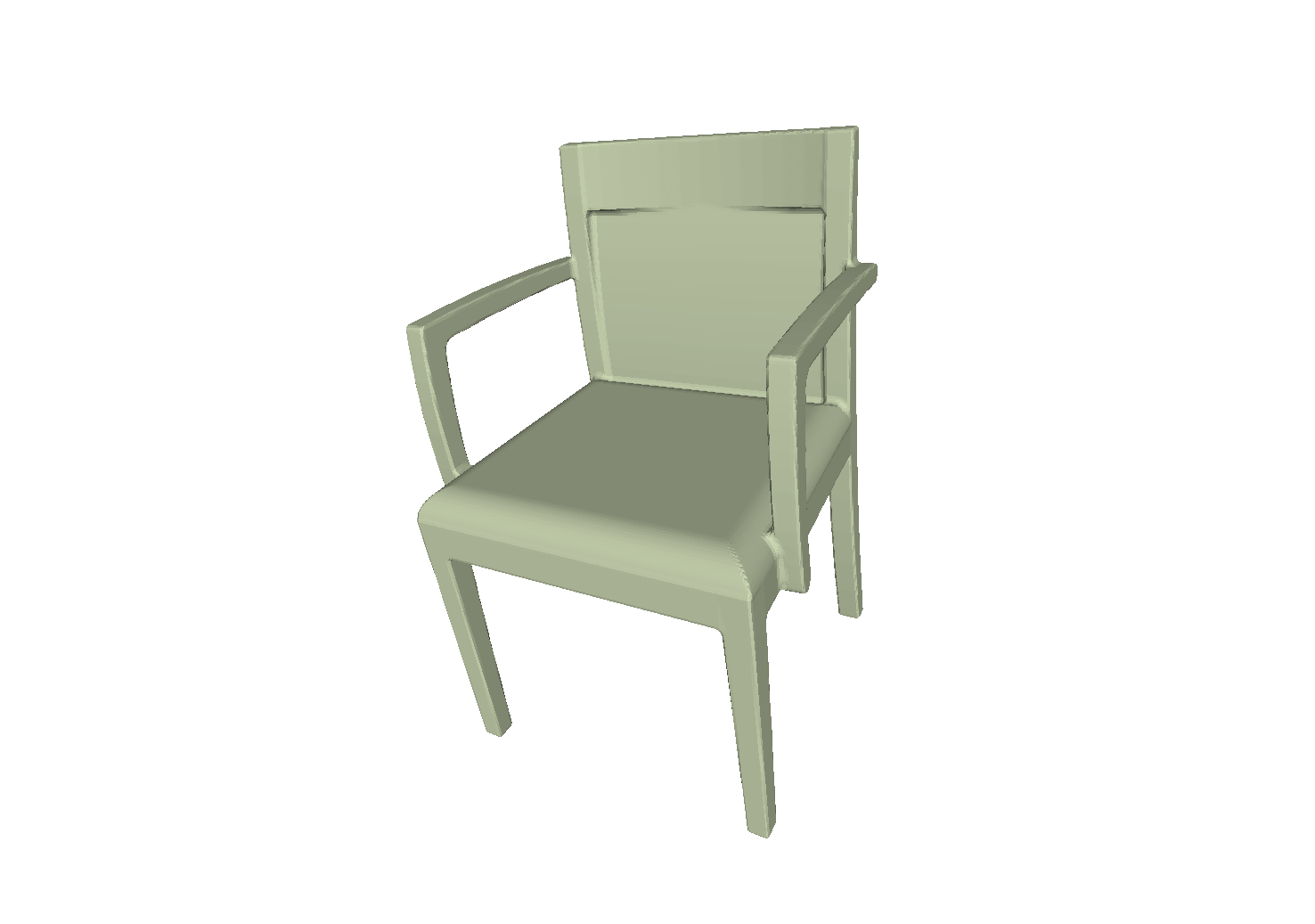}
}
\subfloat{
  \includegraphics[width=0.13\linewidth,height=0.2\linewidth, trim={120 50 0 100},clip]{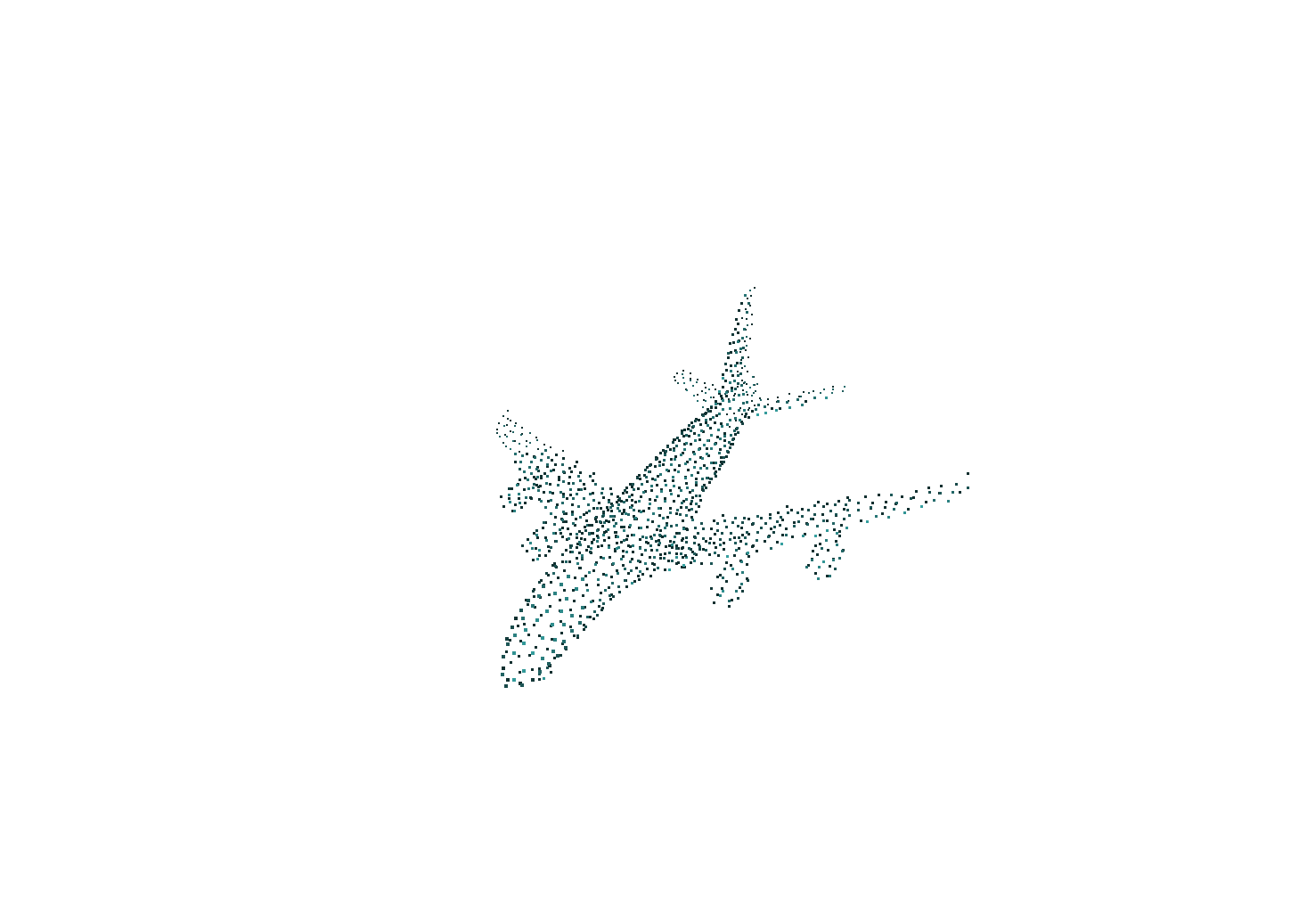} \hspace{-2.2em}
  \includegraphics[width=0.13\linewidth,height=0.2\linewidth, trim={350 50 120 100},clip]{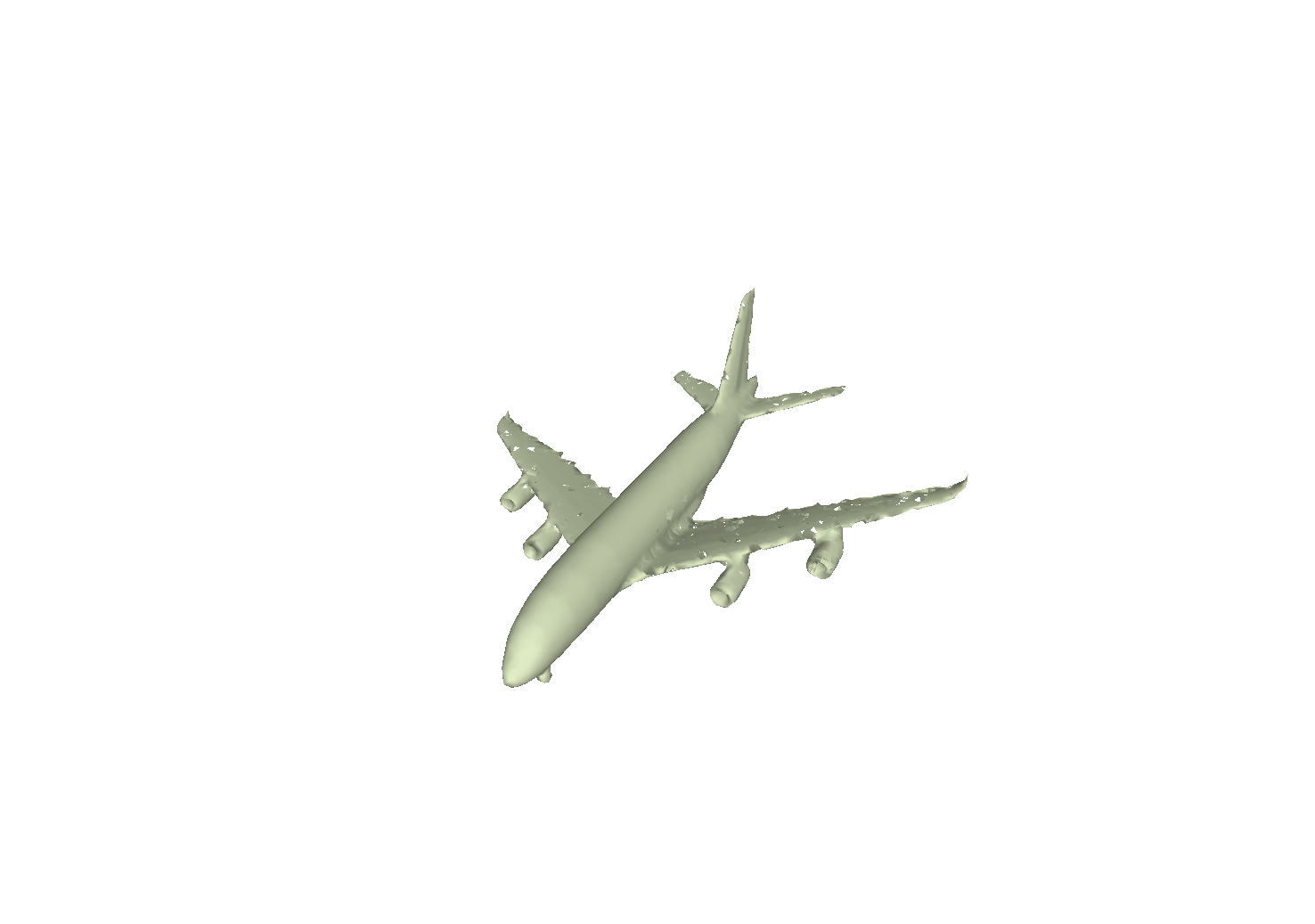} \hspace{-2.2em}
  \includegraphics[width=0.13\linewidth,height=0.2\linewidth, trim={350 50 120 100},clip]{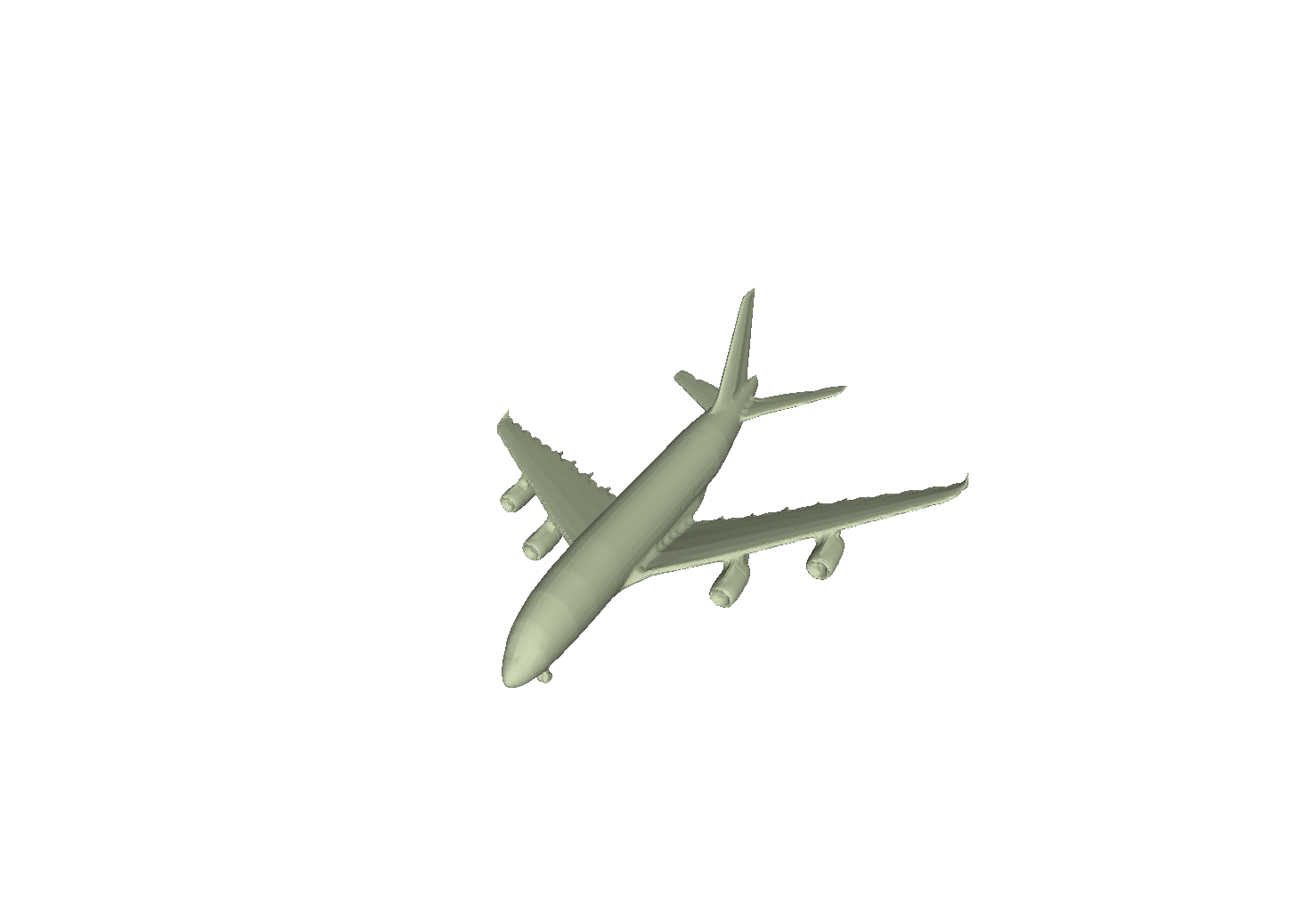}
}
\subfloat{
  \includegraphics[width=0.13\linewidth,height=0.2\linewidth, trim={200 200 200 200},clip]{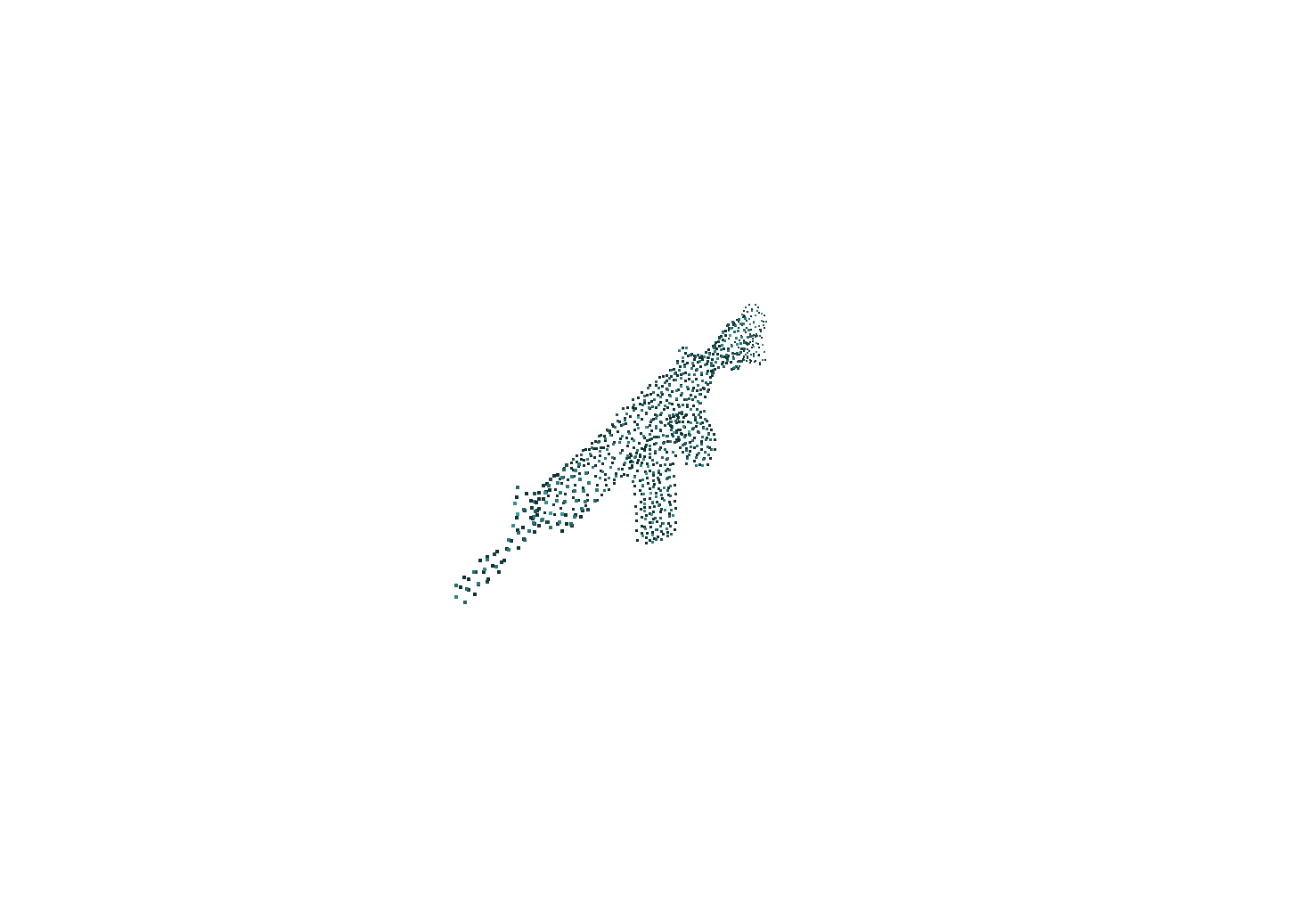} \hspace{-3em}
  \includegraphics[width=0.13\linewidth,height=0.2\linewidth, trim={200 200 200 200},clip]{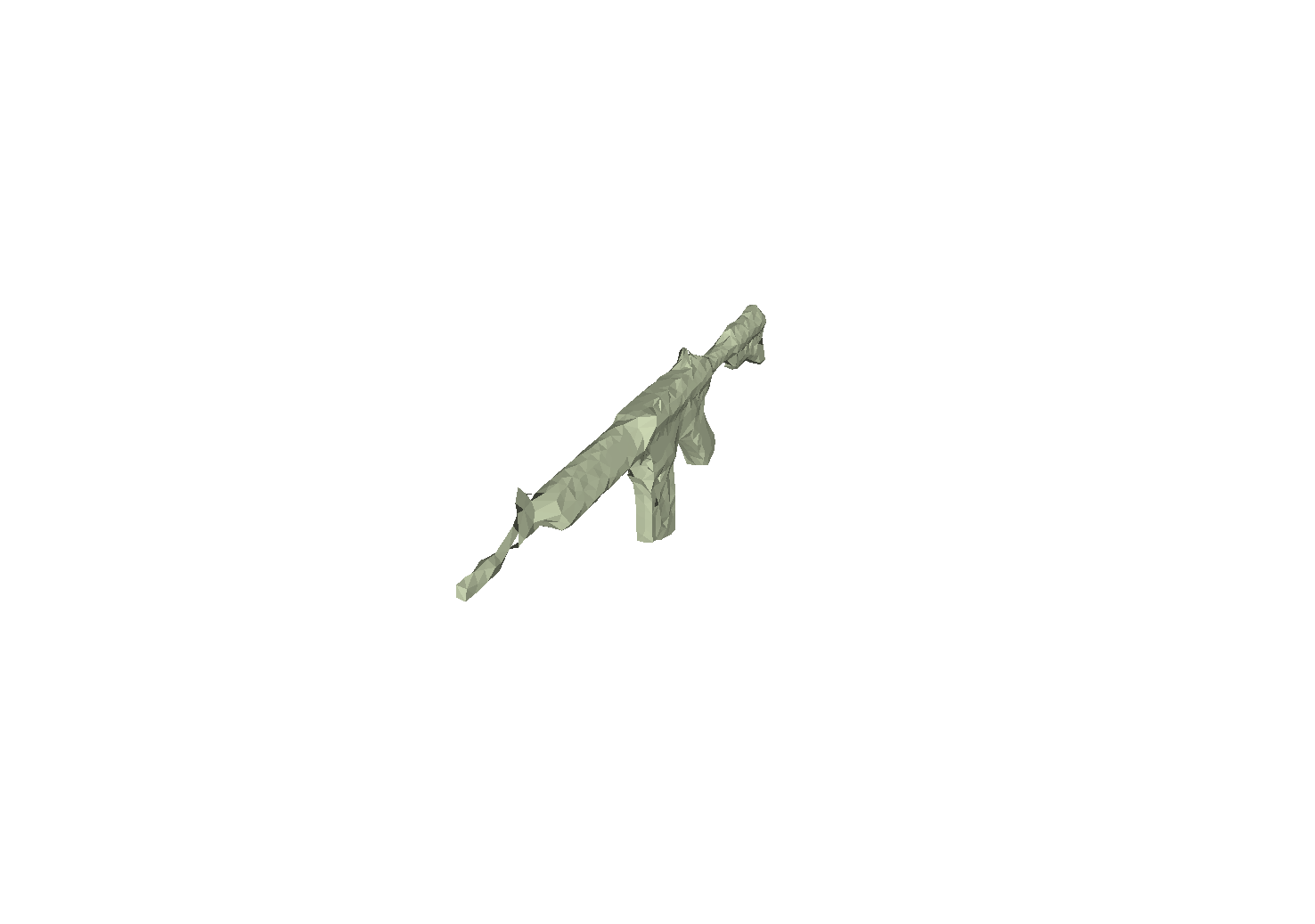} \hspace{-3em}
  \includegraphics[width=0.13\linewidth,height=0.2\linewidth, trim={200 200 200 200},clip]{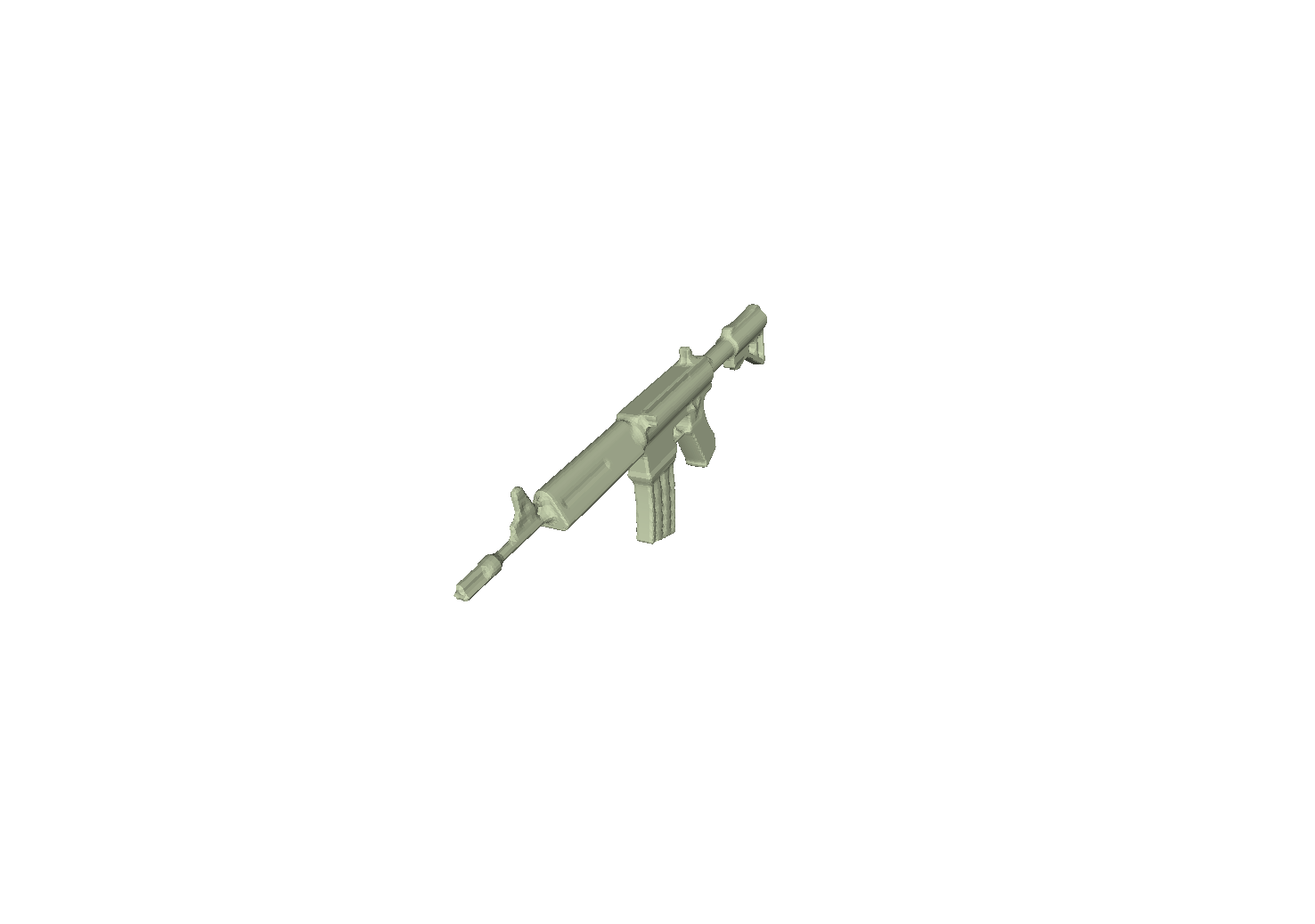}
}
\vspace{-3em} 
\\
\subfloat{
  \includegraphics[width=0.13\linewidth,height=0.2\linewidth, trim={300 0 120 0},clip]{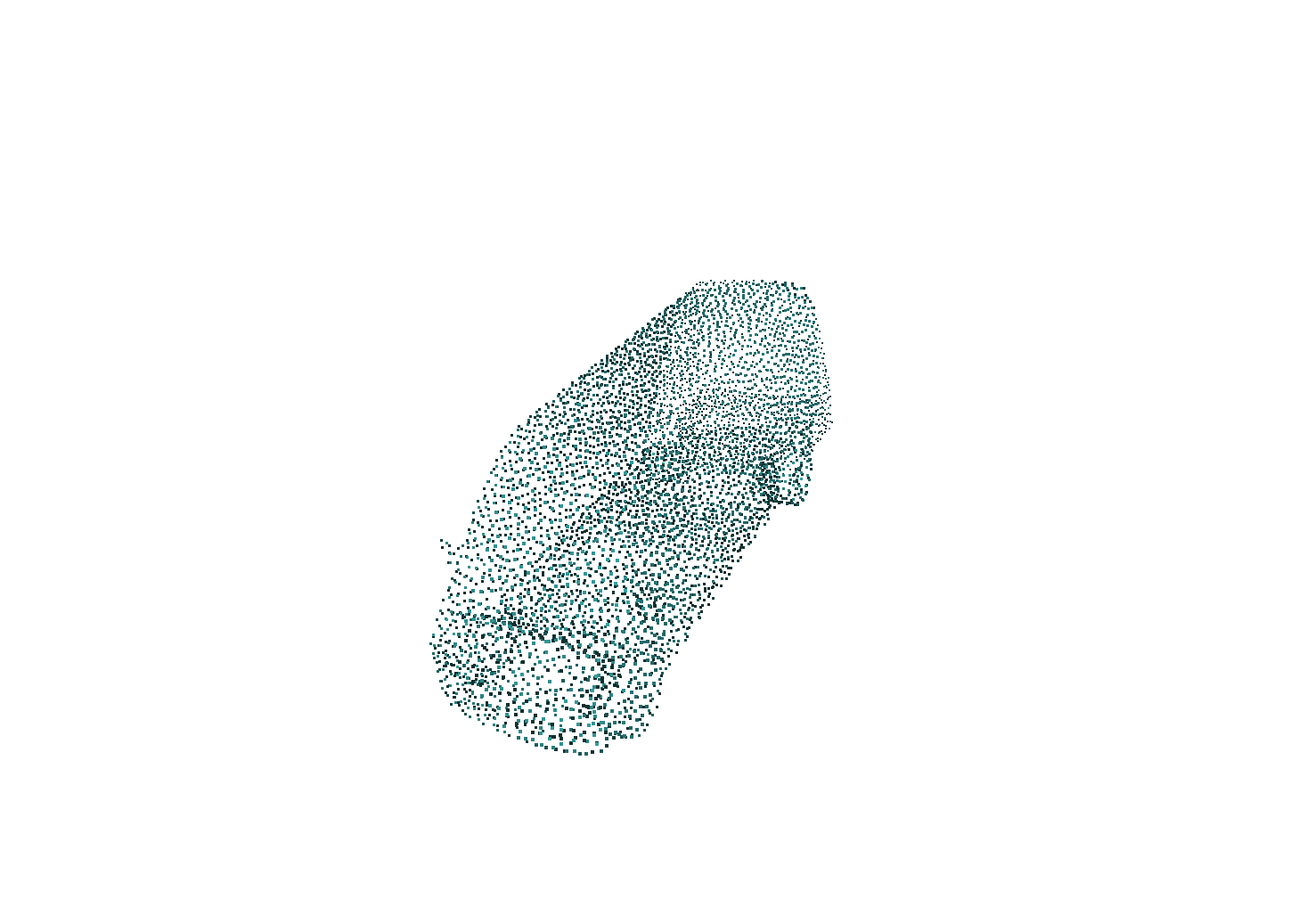} \hspace{-3em}
  \includegraphics[width=0.13\linewidth,height=0.2\linewidth, trim={300 0 120 0},clip]{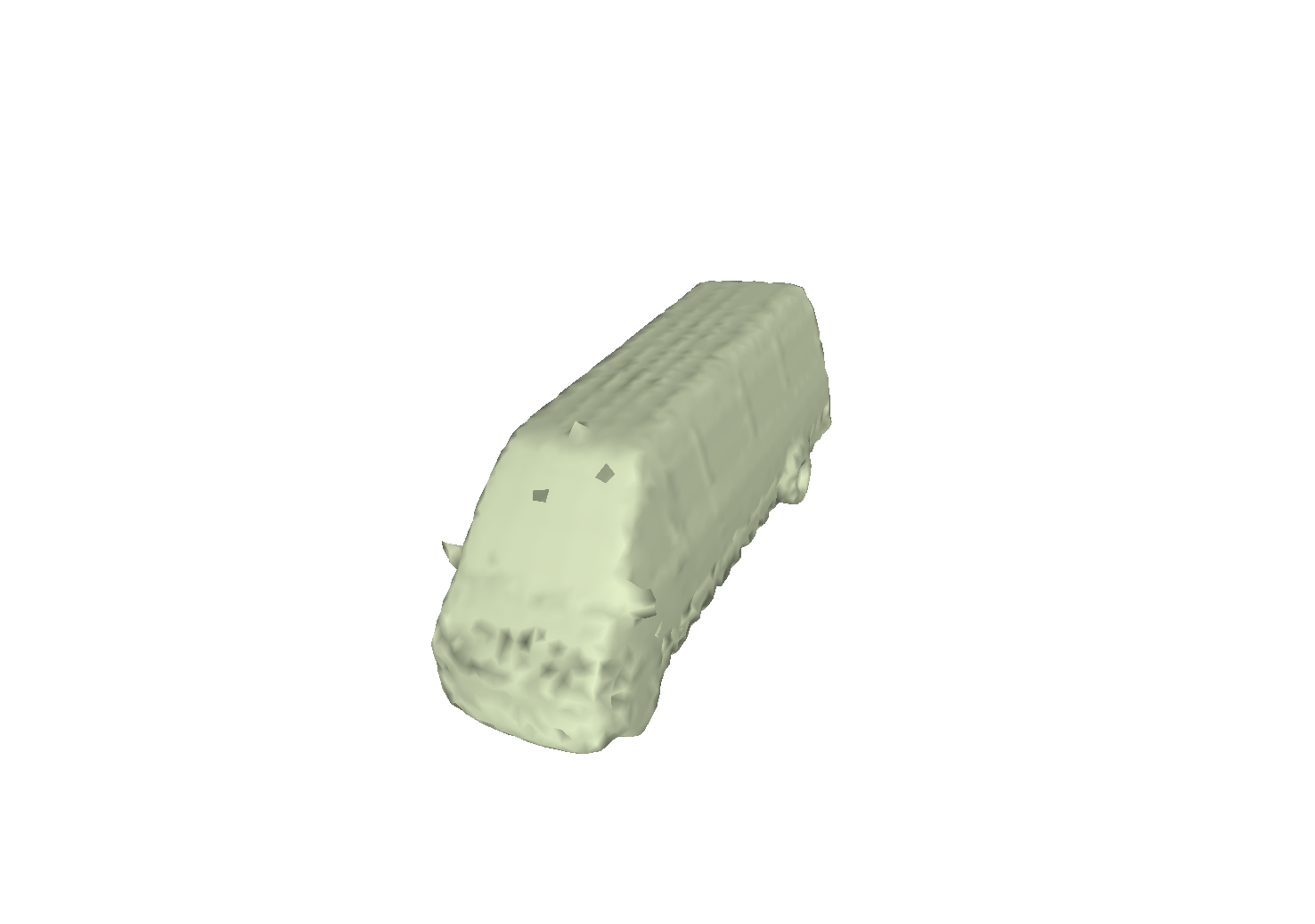} \hspace{-3em}
  \includegraphics[width=0.13\linewidth,height=0.2\linewidth, trim={300 0 120 0},clip]{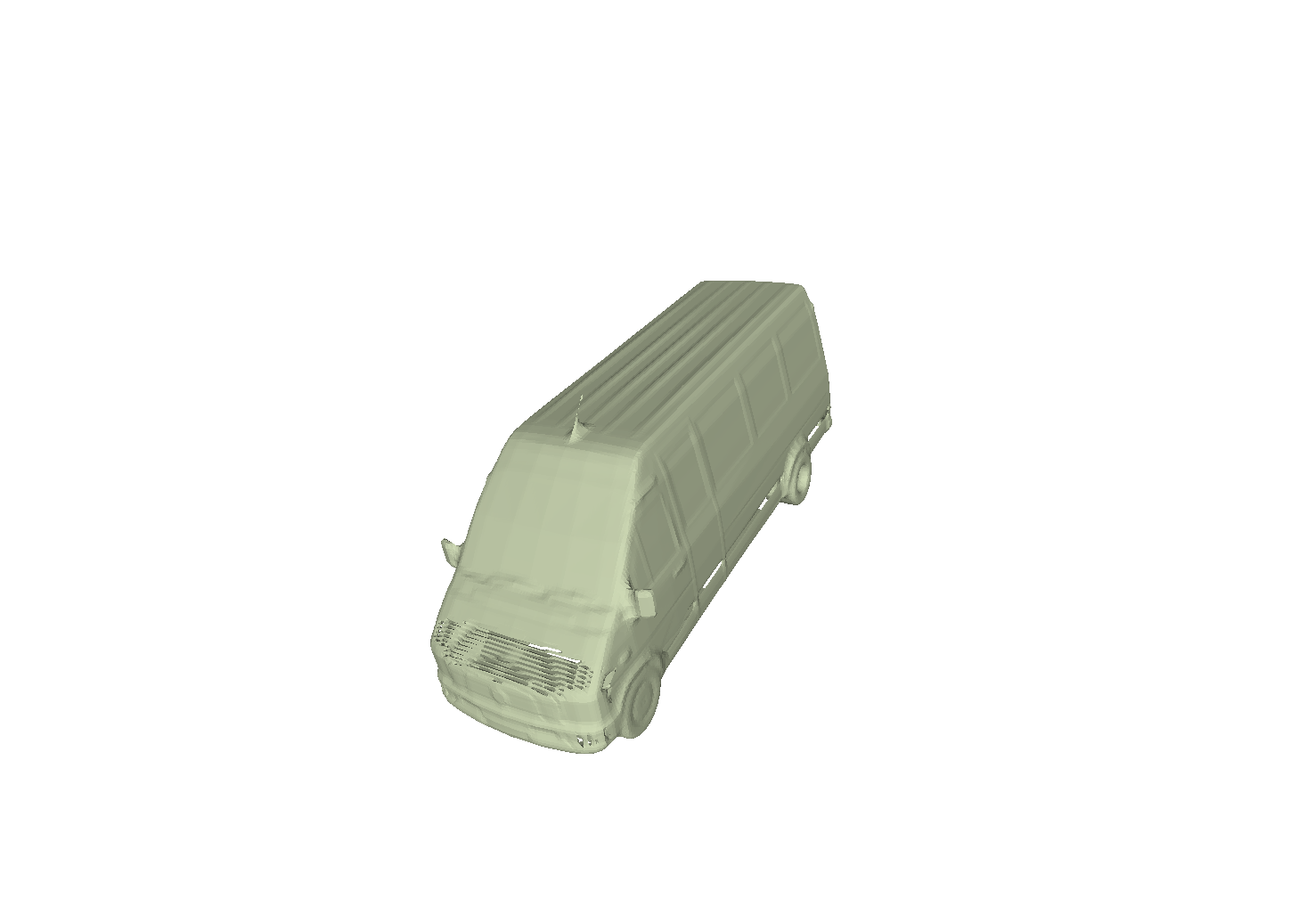}
}
\subfloat{
  \includegraphics[width=0.13\linewidth,height=0.2\linewidth]{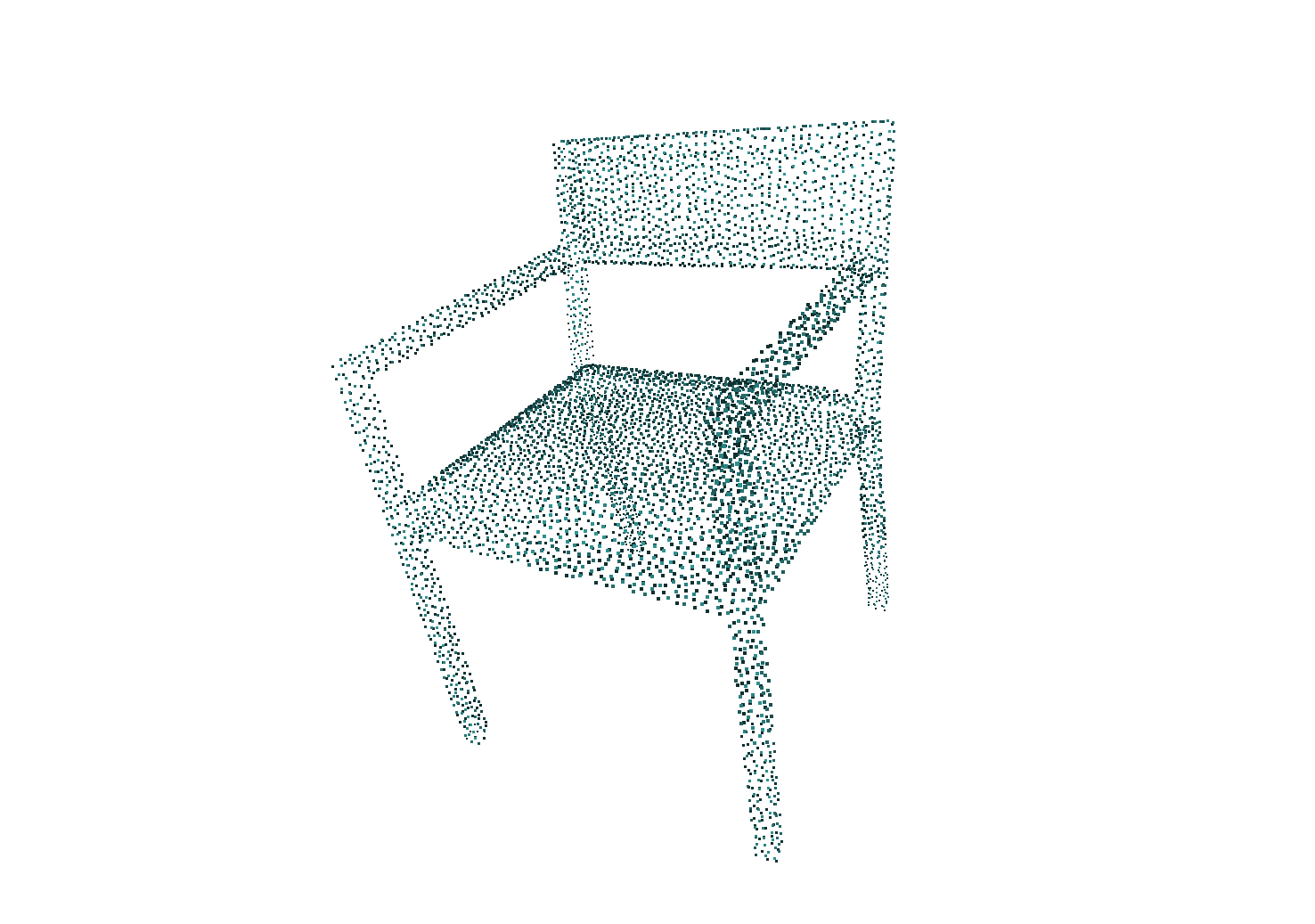} \hspace{-2.5em}
  \includegraphics[width=0.13\linewidth,height=0.2\linewidth]{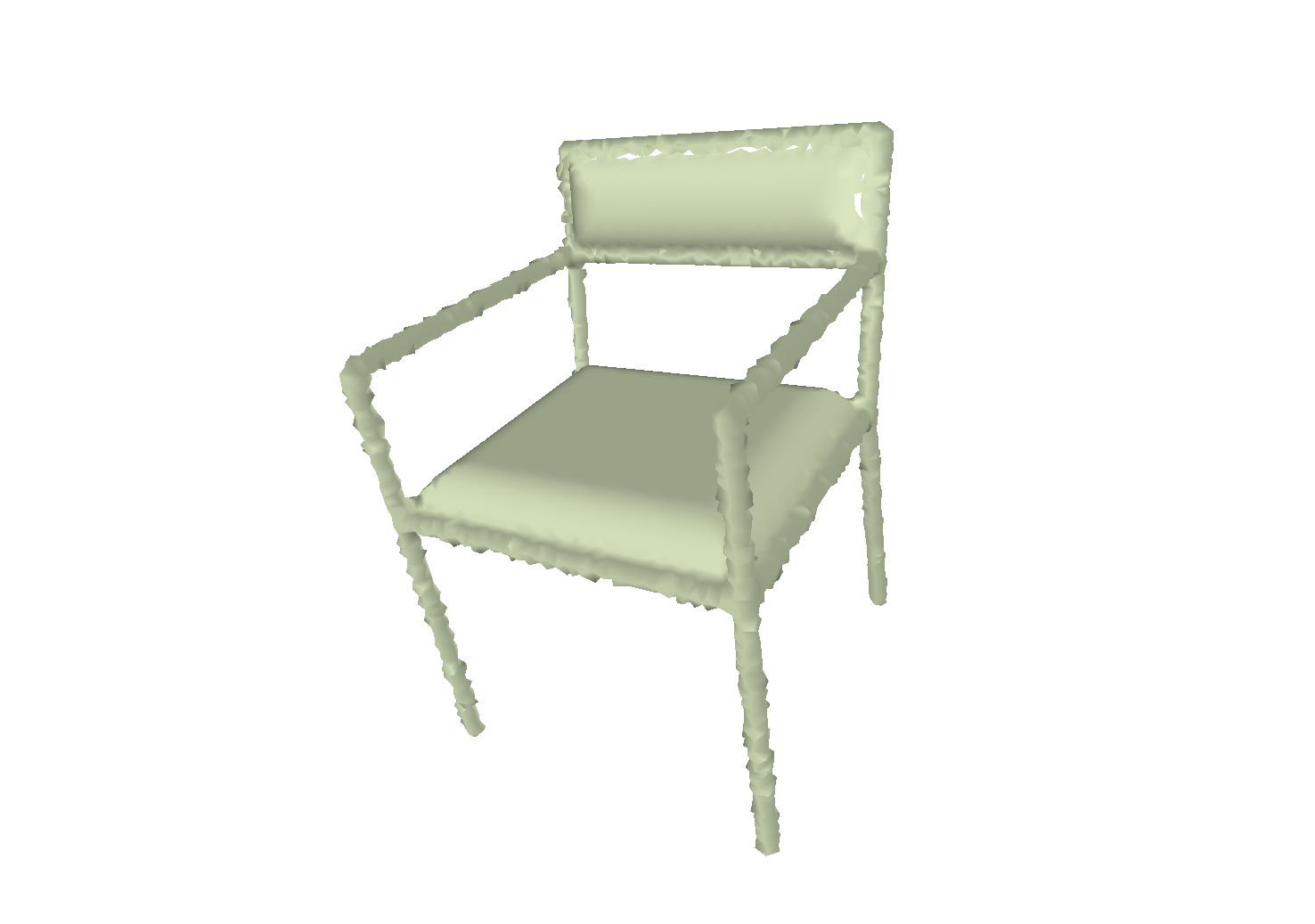} \hspace{-2.5em}
  \includegraphics[width=0.13\linewidth,height=0.2\linewidth]{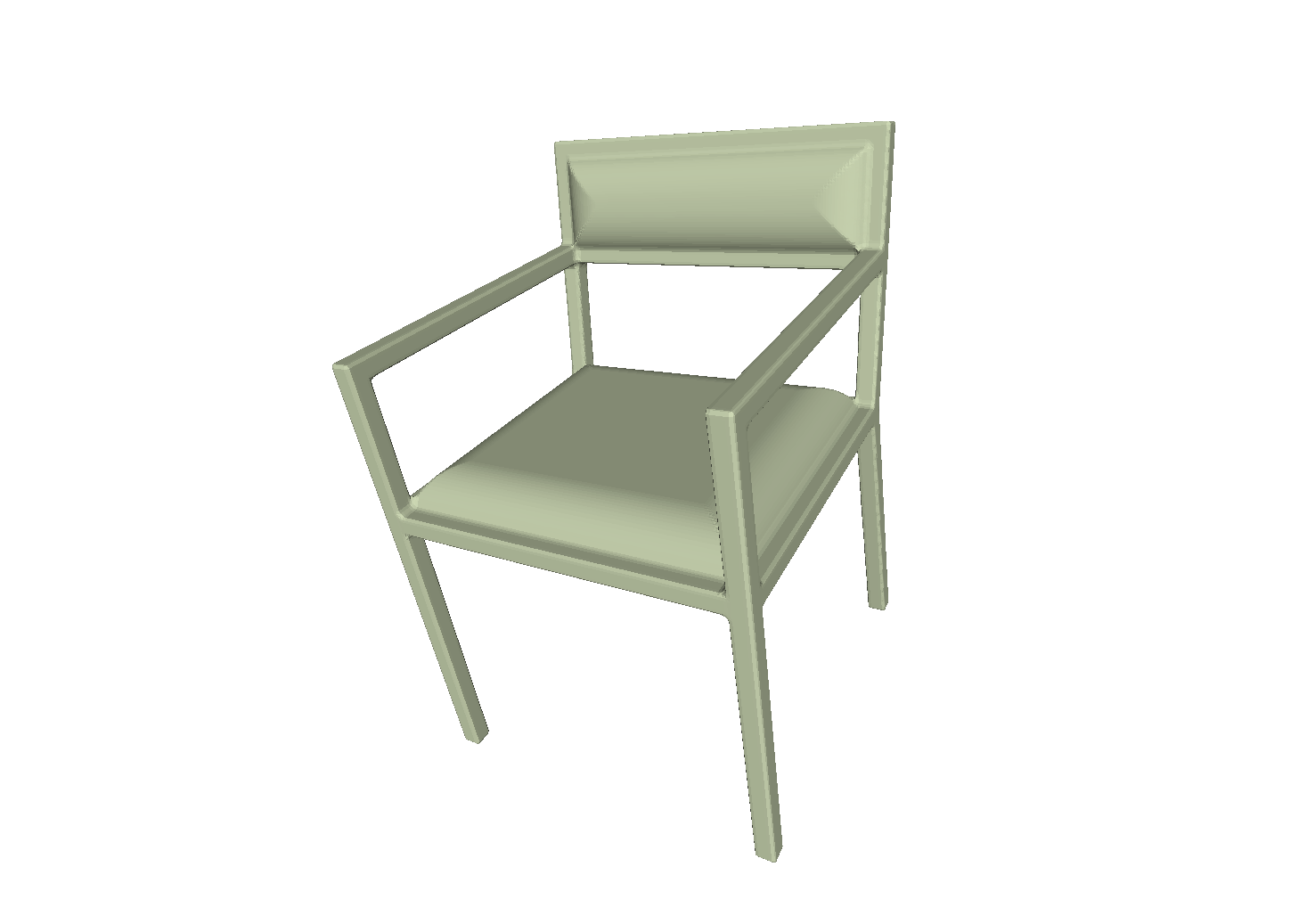}
}
\subfloat{
  \includegraphics[width=0.13\linewidth,height=0.2\linewidth, trim={250 50 120 120},clip]{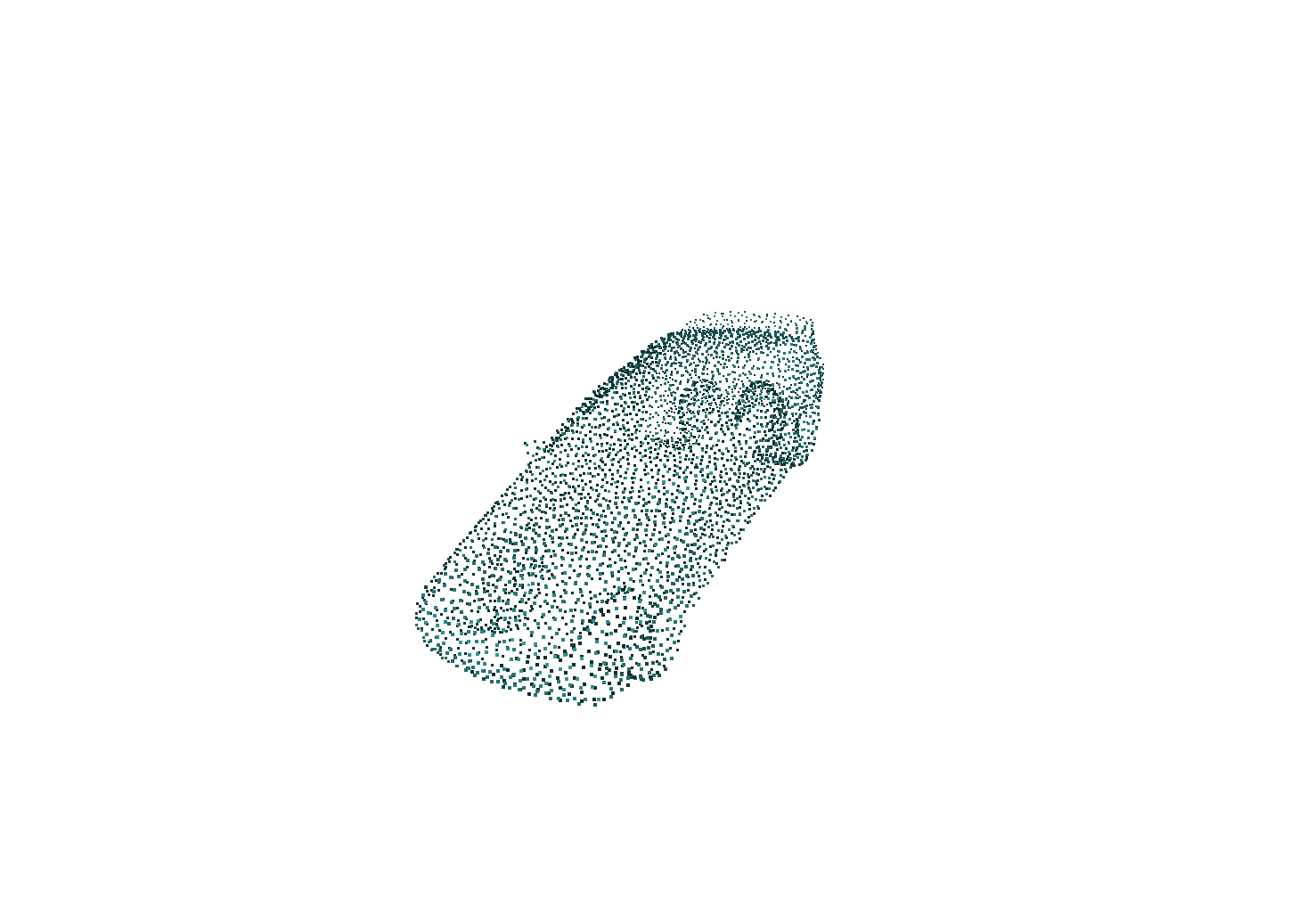} \hspace{-3em}
  \includegraphics[width=0.13\linewidth,height=0.2\linewidth, trim={250 50 120 120},clip]{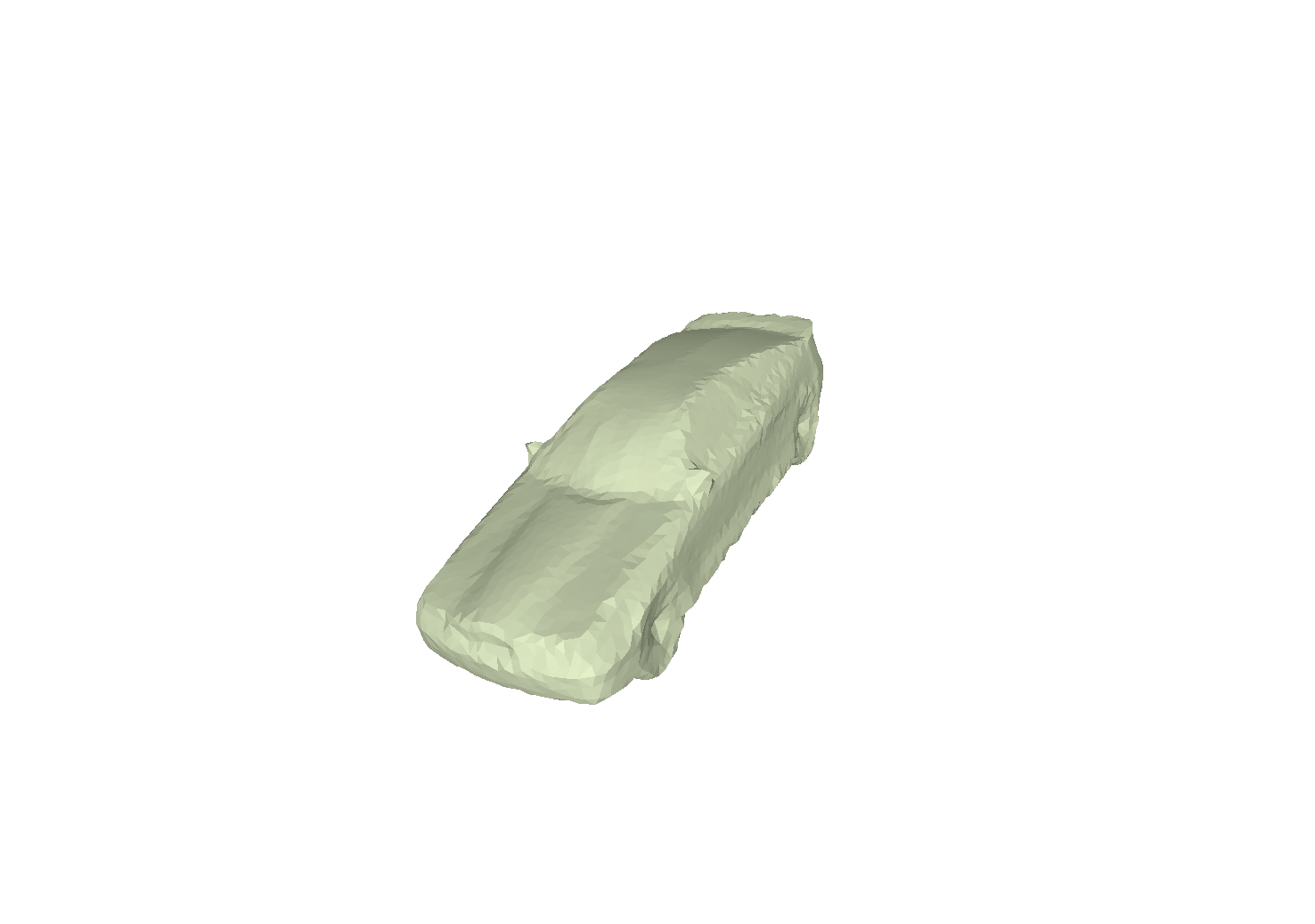} \hspace{-3em}
  \includegraphics[width=0.13\linewidth,height=0.2\linewidth, trim={250 50 120 120},clip]{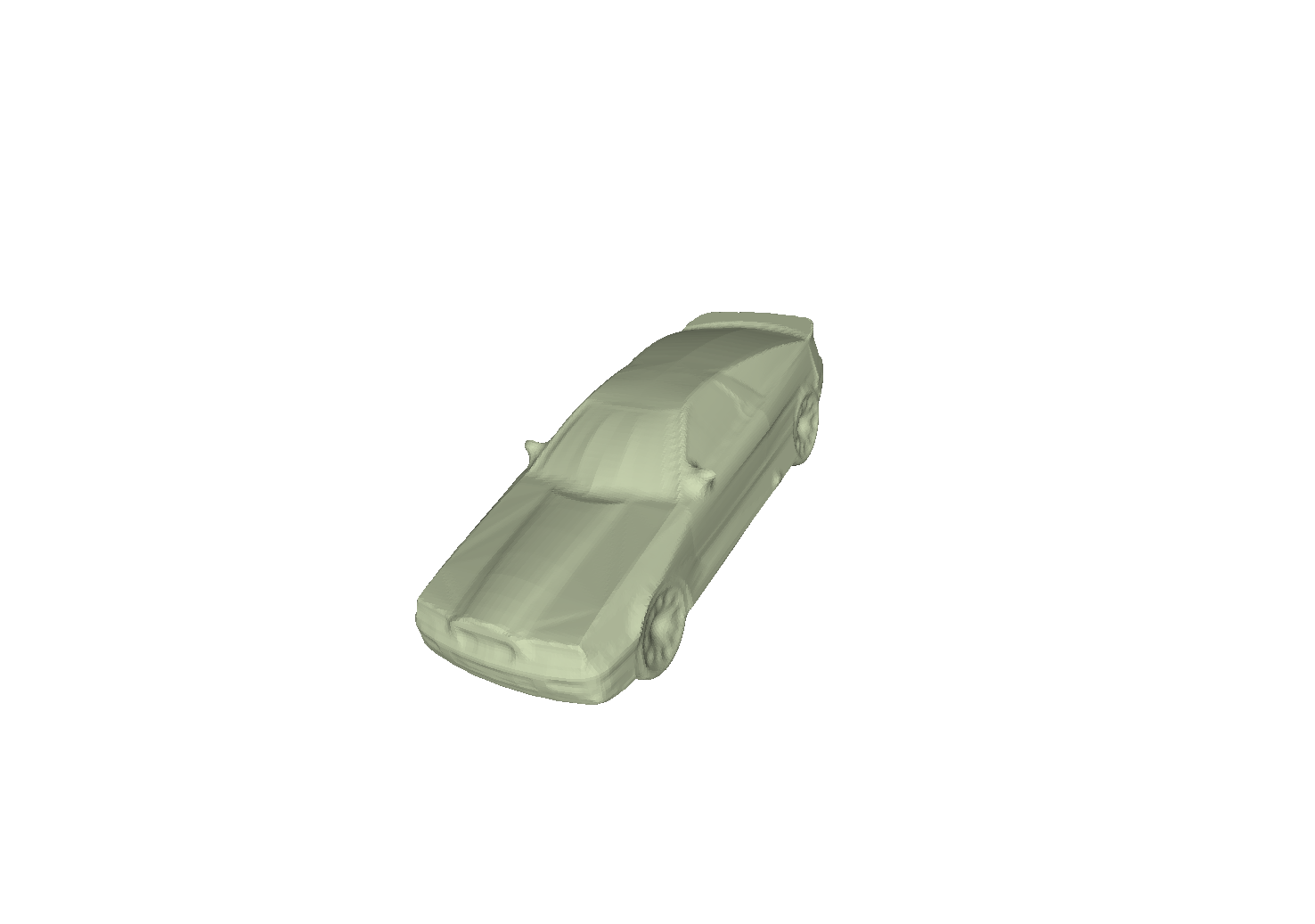}
}
\caption{Reconstruction of closed surfaces from the ShapeNet
\cite{chang2015shapenet} dataset. From left to right, each triplet represents
the input, reconstruction, and ground truth, respectively.}
\label{fig:shapenet_watertight}
\end{figure*}

To understand the effects of sparse input on the reconstruction quality, we
evaluated \modelname\ and the baseline using an input density of $N \sim
\{300,3000,10000\}$ points while fixing the voxel resolution to $M = 256$. In
contrast to the baseline, \modelname\ can reconstruct thin structures more
accurately and preserve small gaps (Fig.~\ref{fig:result_obj_recon} inset
images) while quantitatively outperforming the reconstruction with different
input densities (Table~\ref{tab:result_obj_recon}). Furthermore, we
investigated \modelname's ability to perform closed-surface reconstruction on
preprocessed watertight meshes using 13 subsets of ShapeNet for training. The
reconstruction results are shown in Fig.~\ref{fig:shapenet_watertight}.

\begin{table}
\begin{tabularx}{\columnwidth}{c | Y | Y | Y | Y | Y } \hline
              & \multicolumn{3}{c|}{\textit{Chamfer-}$L_2 \downarrow$} &  
              \multicolumn{2}{c}{$F$-\textit{score} $\uparrow$}\\ 
              \cline{2-6}
              & $ N = 300$          & $N = 3000$        & $N = 10000$ 
              & $d = 0.1\%$         & $d = 0.05\%$ \\ \hline
  NDF         & $1.550$             & $0.324$           & $0.092$ 
              & $0.711$             & $0.460$ \\
  \modelname\ & $\mathbf{1.217}$    & $\mathbf{0.119}$  & $\mathbf{0.068}$ 
              & $\mathbf{0.785}$    & $\mathbf{0.542}$ \\ \hline
\end{tabularx}
\caption{A quantitative comparison between \modelname\ and NDF
\cite{chibane2020neural} on the ShapeNet Cars \cite{chang2015shapenet} dataset
for object reconstruction from different input densities. \modelname\
outperforms NDF on all input densities. The chamfer-$L_2$ results are of order
$\times 10^{-4}$ and the reconstruction results using an input density of $N =
10000$ were used to calculate the F-score.}
\label{tab:result_obj_recon}
\end{table}

\begin{table}
\begin{tabularx}{\columnwidth}{c | Y | Y | Y | Y | Y | Y} \hline
              & \multicolumn{3}{c|}{\textit{Chamfer-}$L_2 \downarrow$} &  
              \multicolumn{3}{c}{$F$-\textit{score}$_{0.05}$ $\uparrow$}\\ 
              \cline{2-7}
              & $64^3$              & $128^3$           & $256^3$ 
              & $64^3$              & $128^3$           & $256^3$ \\ \hline
NDF           & $1.549$             & $0.266$           & $0.029$ 
              & $0.289$             & $0.591$           & $0.994$\\
GIFS          & $5.245$             & $1.210$           & $0.141$ 
              & $0.240$             & $0.510$           & $0.891$\\
\modelname\   & $\mathbf{1.441}$    & $\mathbf{0.162}$  & $\mathbf{0.023}$ 
              & $\mathbf{0.335}$    & $\mathbf{0.803}$  & $\mathbf{0.995}$\\ \hline
\end{tabularx}
\caption{A quantitative comparison between NDF \cite{chibane2020neural}, GIFS
\cite{ye2022gifs}, and \modelname~on the Garments \cite{bhatnagar2019multi}
dataset for object reconstruction at different voxel resolutions. \modelname\
outperforms the baselines by significant margin in lower resolutions. The point
density was fixed to $N=3K$ for this experiment. The chamfer-$L_2$ results
are of order $\times 10^{-4}$.}
\label{tab:garments_reconstruction}
\end{table}

\subsection{Real-world scene reconstruction}
\label{sec:real-world_scene_reconstruction}
\begin{figure*}
\centering
\subfloat[Input]
  {\includegraphics[width=0.15\linewidth,height=3cm]{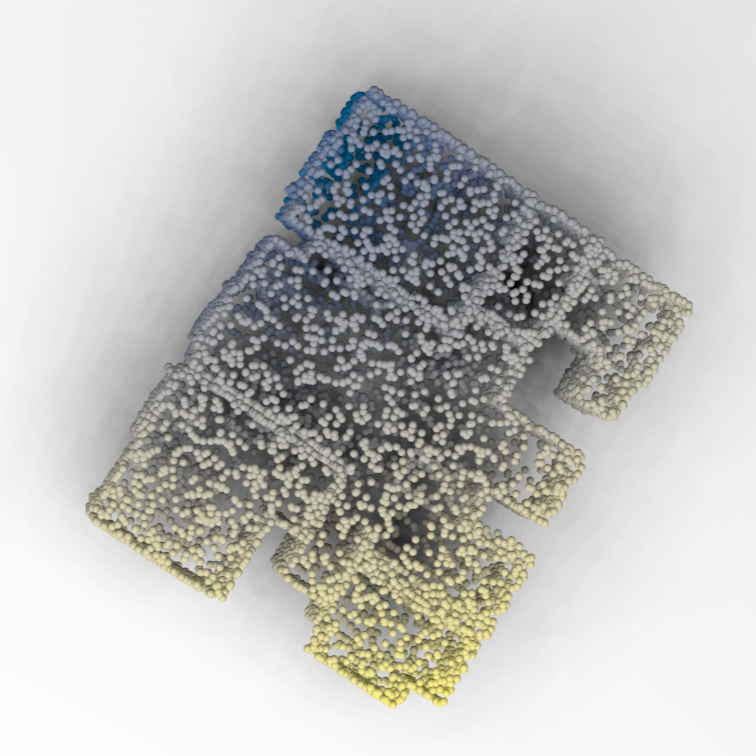}} \hspace{0.05pt}
\subfloat[NDF]
  {\includegraphics[width=0.15\linewidth,height=3cm]{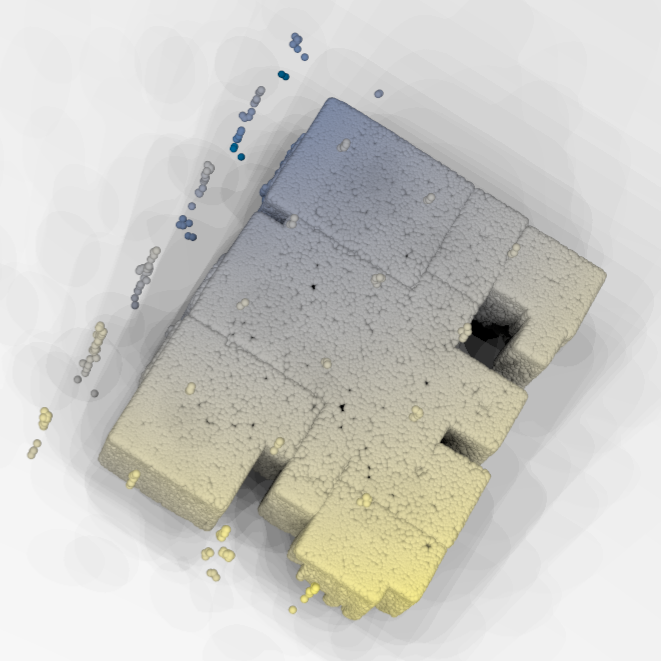}} \hspace{0.05pt}
\subfloat[\modelname]
  {\includegraphics[width=0.15\linewidth,height=3cm]{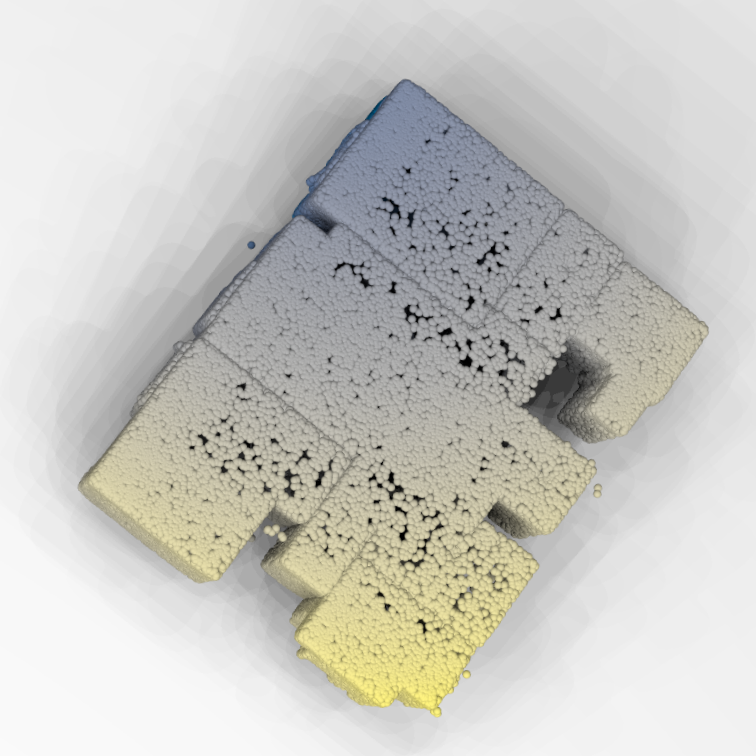}} \hspace{0.05pt}
\subfloat[\modelname$_\text{mesh}$]
   {\includegraphics[width=0.15\linewidth,height=2.5cm]{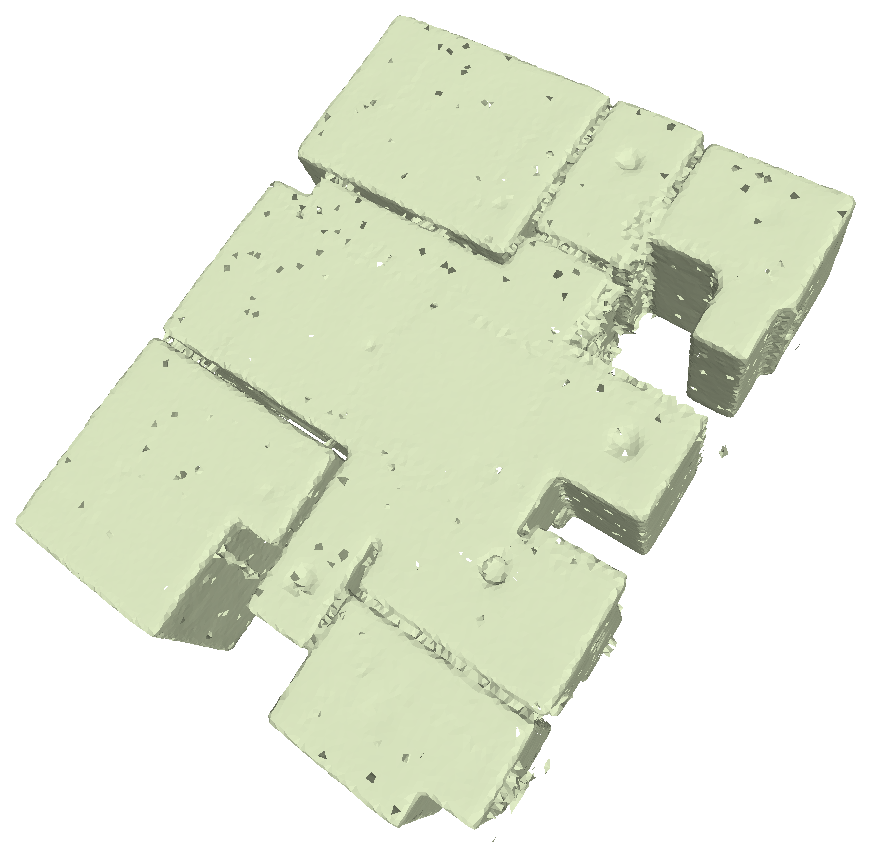}} \hspace{0.05pt}
\subfloat[GT]
  {\includegraphics[width=0.15\linewidth,height=3cm]{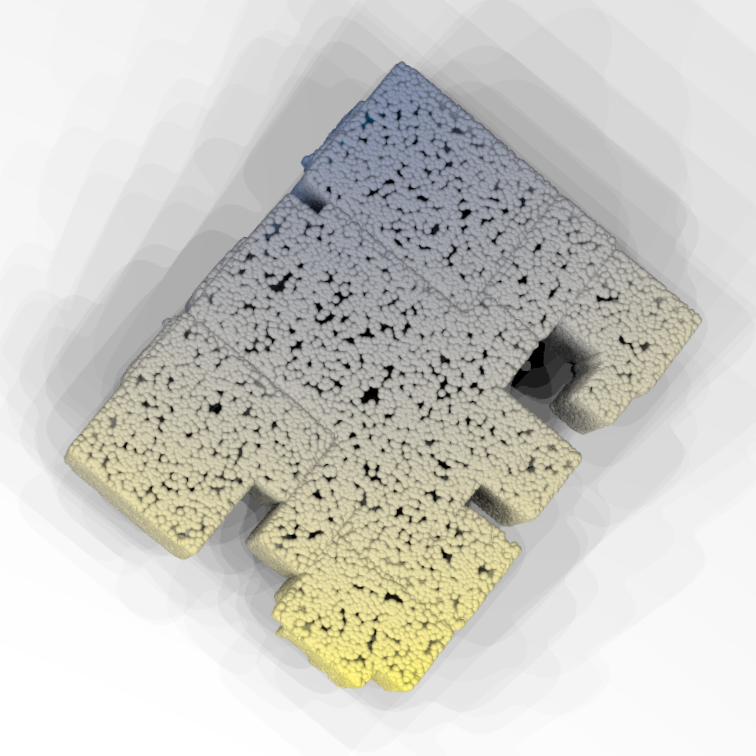}}
\par \vspace{-0.9em}
\subfloat
  {\includegraphics[width=0.15\linewidth,height=3cm]{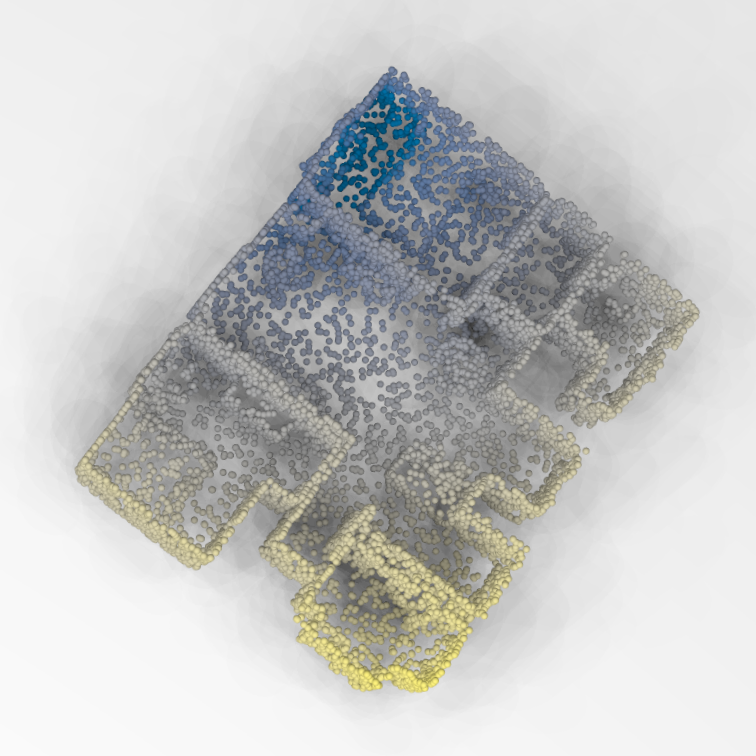}} \hspace{0.05pt}
\subfloat
  {\includegraphics[width=0.15\linewidth,height=3cm]{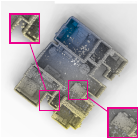}} \hspace{0.05pt}
\subfloat
  {\includegraphics[width=0.15\linewidth,height=3cm]{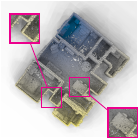}} \hspace{0.05pt}
\subfloat
  {\includegraphics[width=0.15\linewidth,height=2.5cm]{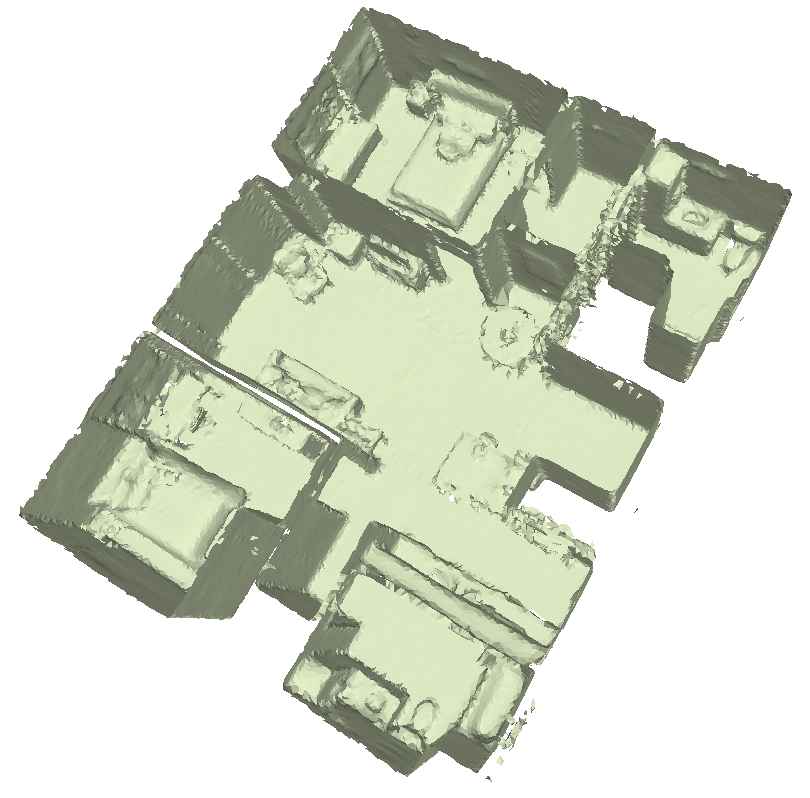}} 
  \hspace{0.05pt}
\subfloat
  {\includegraphics[width=0.15\linewidth,height=4cm]{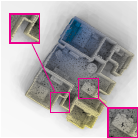}} \hspace{0.05pt}
\par \vspace{-0.9em}
\subfloat
  {\includegraphics[width=0.15\linewidth,height=3cm]{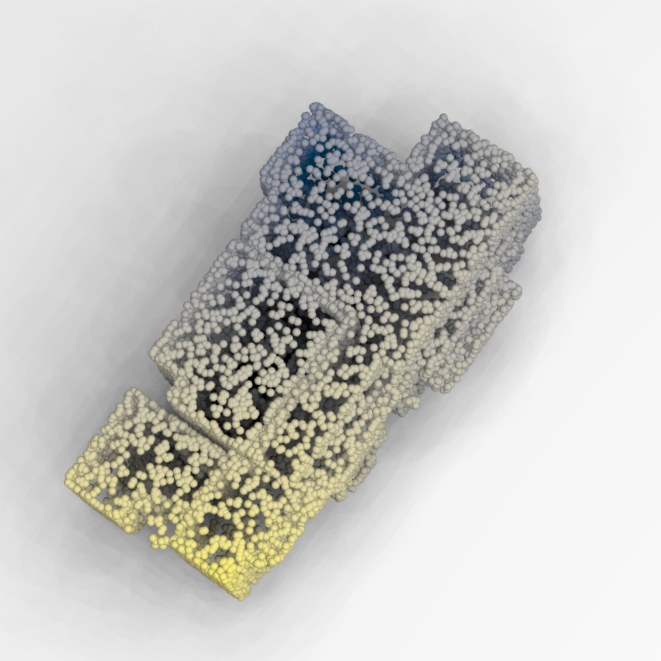}} 
  \hspace{0.05pt}
\subfloat
  {\includegraphics[width=0.15\linewidth,height=3cm]{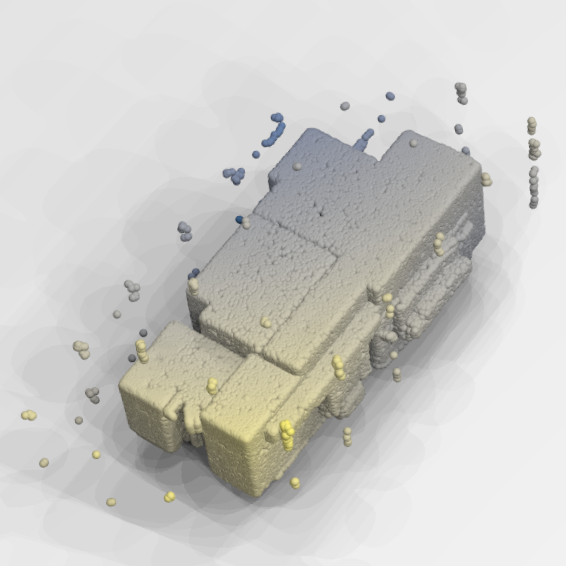}} 
  \hspace{0.05pt}
\subfloat
  {\includegraphics[width=0.15\linewidth,height=3cm]{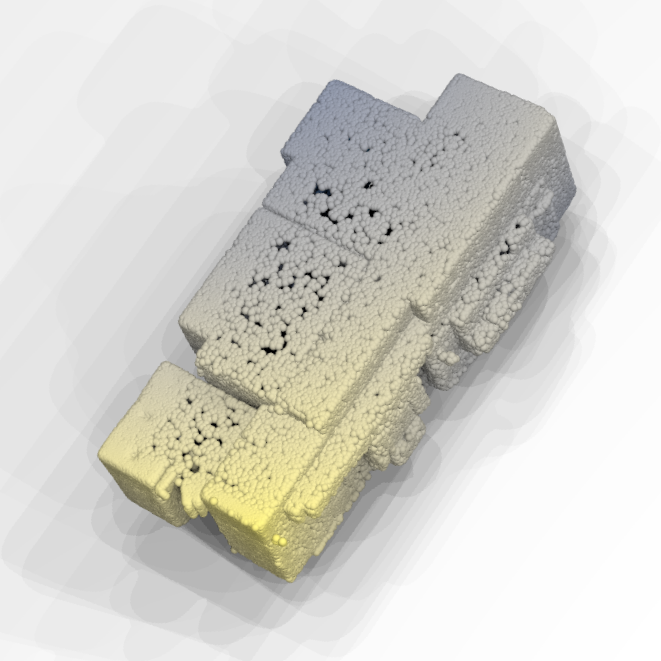}} 
  \hspace{0.05pt}
\subfloat
  {\includegraphics[width=0.15\linewidth,height=2.5cm]{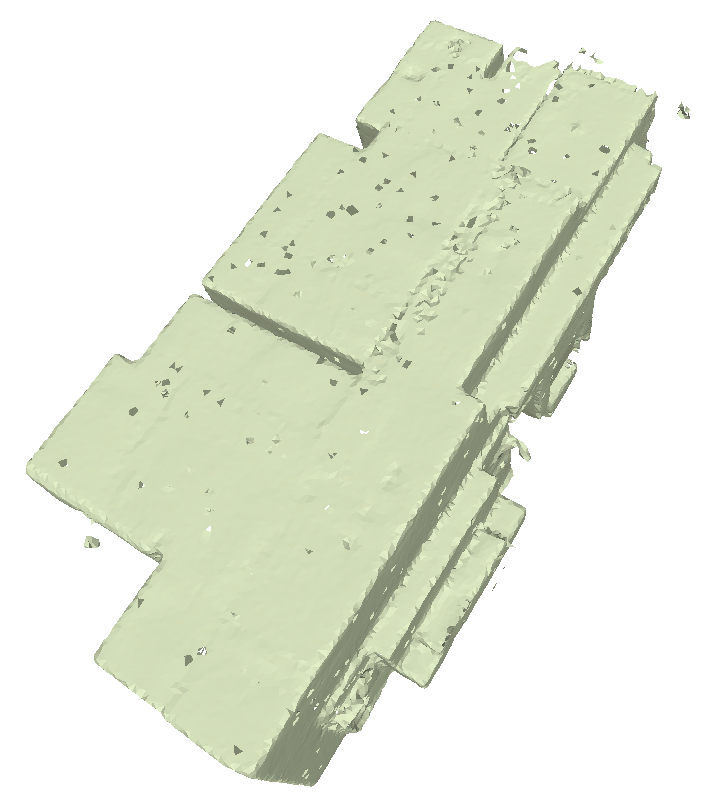}} \hspace{0.05pt}
\subfloat
  {\includegraphics[width=0.15\linewidth,height=3cm]{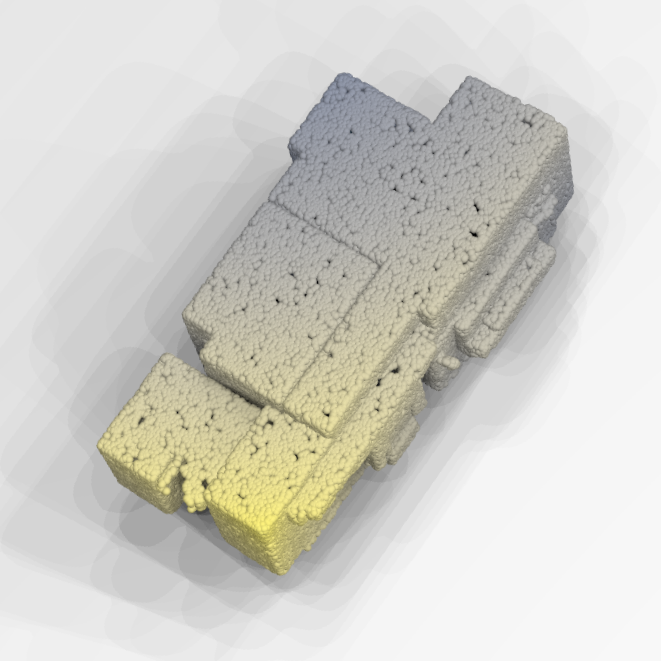}}
\par \vspace{-0.9em}
\subfloat
  {\includegraphics[width=0.15\linewidth,height=3cm]{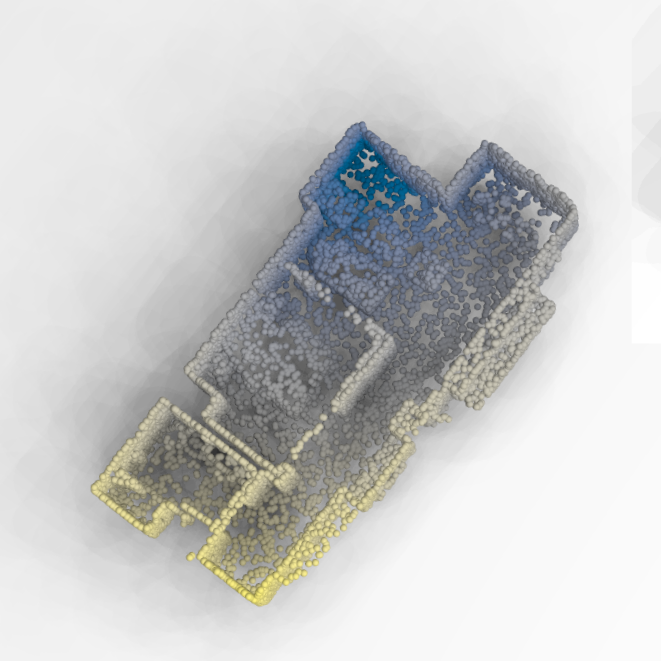}} \hspace{0.05pt}
\subfloat
  {\includegraphics[width=0.15\linewidth,height=3cm]{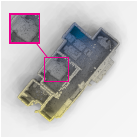}} \hspace{0.05pt}
\subfloat
  {\includegraphics[width=0.15\linewidth,height=3cm]{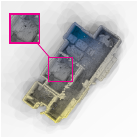}} \hspace{0.05pt}
\subfloat
  {\includegraphics[width=0.15\linewidth,height=2.5cm]{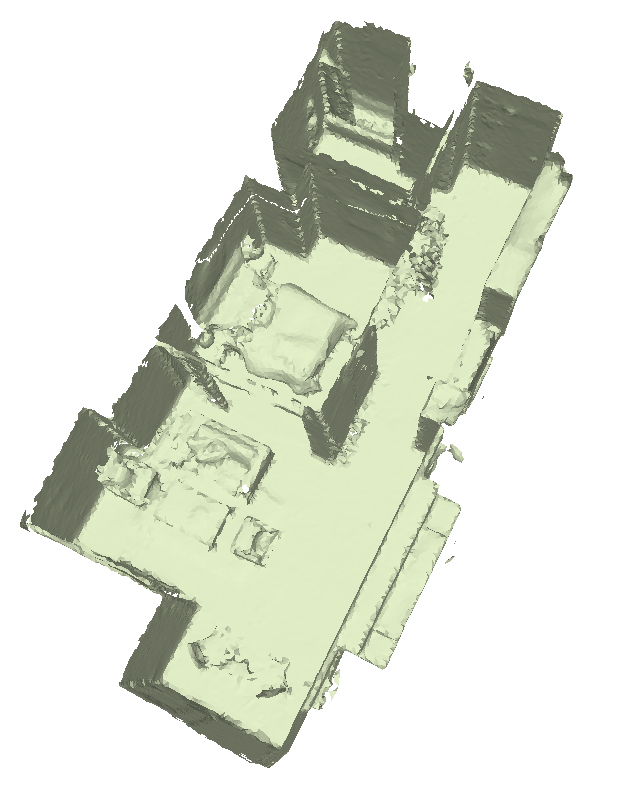}} 
  \hspace{0.05pt}
\subfloat
  {\includegraphics[width=0.15\linewidth,height=3cm]{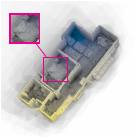}} \hspace{0.05pt}
\par \vspace{-0.9em}
\caption{Scene reconstruction on the test set of the Gibson Environment
\cite{xia2018gibson} dataset using NDF \cite{chibane2020neural}, \modelname, and
the respective ground truth (GT). Each odd row represents an outside view of a
scene while the even rows depict inside views. In contrast to the baseline,
\modelname\ produces significantly less outliers (outside view) and improves the
preservation of geometric features (inset images).} 
\label{fig:result_scene_recon}
\end{figure*}

\begin{figure*}
\centering
\includegraphics[scale=0.32]{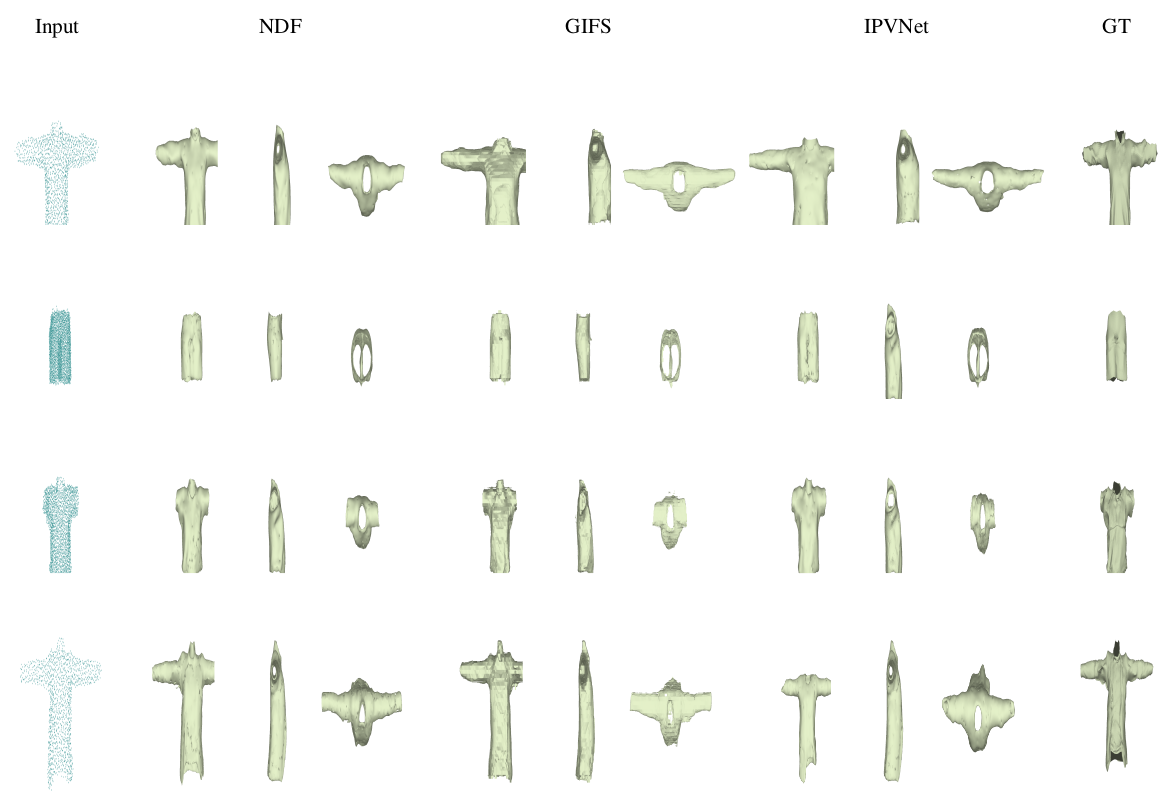}
\caption{A qualitative comparison between NDF \cite{chibane2020neural}, GIFS
\cite{ye2022gifs}, and \modelname\ on the Garments \cite{bhatnagar2019multi}
dataset.}
\label{fig:garments_reconstruction}
\end{figure*}

We evaluated the reconstruction of complex real-world scenes through the use of
the Gibson Environment dataset \cite{xia2018gibson}. The dataset consists of
RGBD scans of indoor spaces. A subset of 35 and 100 scenes were prepared
following the procedure from \cite{chibane2020neural} for training and testing,
respectively. We utilized a sliding window scheme and reconstructed the surface
bounded by each window. Since the sliding window may frequently consist of a
very small area of the scene with only few points, we used an output density
five times as large the input density (i.e., $\mathcal{O} = 5 \times N$) to save
time. The grid resolutions were kept fixed at $M = 256$ for both \modelname\ and
the baseline. The reconstruction results are highlighted in
Fig.~\ref{fig:result_scene_recon}. In addition to improving the preservation of
structural details, \modelname\ produces significantly fewer outliers than the
baseline due to the use of point features during training and inference.

Lastly, we tested \modelname\ on the challenging complex surfaces of the
Garments \cite{bhatnagar2019multi} dataset.
Fig.~\ref{fig:garments_reconstruction} and
Table~\ref{tab:garments_reconstruction} show the qualitative and quantitative
results, respectively. It can be observed from
Table~\ref{tab:garments_reconstruction} that \modelname\ exhibits superior
performance compared to the baselines, particularly at low grid resolutions.
The point-voxel fusion technique utilized by our model is able to effectively
recover lost details from the resulting discretization.

\section{Ablation study}
\label{sec:ablation_study}
In this section, we study the effect of different design choices and how they
influence the performance of \modelname\ on the task of 3D reconstruction.

\subsection{Effect of point features on object reconstruction}
\label{subsec:effect_of_point_features_on_object_reconstruction}
Since multiple points within the boundary of a grid are merged together in low
resolutions, we test the effect of this information loss on object
reconstruction. To understand if the point features are helpful in recovering
missing information, we trained a version of \modelname\ named
\modelname$_{\text{wp}}$, which has the same neural functions except for the
point encoder and point-feature aggregation. Both \modelname\ and
\modelname$_{\text{wp}}$ were trained with differing grid resolutions, $M \sim
\{32,64,128,256\}$, while using a fixed input point density of $N = 10000$. Our
findings on the reconstruction of the ShapeNet Cars dataset are illustrated in
Table~\ref{tab:vox_vs_point_voxel}. At lower resolutions, where a significant
percentage of the raw points are lost due to voxelization, \modelname\
outperforms \modelname$_{\text{wp}}$ by a notable margin thus indicating the
usefulness of point features.

\begin{table}
\begin{tabularx}{\columnwidth}{Y | Y | Y | Y } \hline
  \multirow{2}{*}{\shortstack{Grid \\ Resolution}}        &
  \multirow{2}{*}{\shortstack{$\%$ of \\ Lost Points}}  &
  \multicolumn{2}{c}{\textit{Chamfer-}$L_2 \downarrow$} \\ \cline{3-4}
        &        & \modelname$_{\text{wp}}$  & \modelname\ \\ \hline
   32   & $82\%$ & $9.587$        & $\mathbf{4.307}$ \\
   64   & $45\%$ & $0.961$        & $\mathbf{0.543}$ \\
   128  & $16\%$ & $0.395$        & $\mathbf{0.257}$ \\
   256  & $4\%$  & $0.092$        & $\mathbf{0.068}$ \\\hline
\end{tabularx}
\caption{The object reconstruction accuracy for different grid resolutions on
the ShapeNet Cars \cite{chang2015shapenet} dataset using only voxel features
(\modelname$_{\text{wp}}$) and point-voxel features (\modelname). The second
column represents the percentage of raw points lost during the voxelization
process due to multiple points overlapping in the same grid. In low-resolution
grids, \modelname\ significantly outperforms \modelname$_{\text{wp}}$. The
chamfer-$L_2$ results are of order $\times 10^{-4}$.}
\label{tab:vox_vs_point_voxel}
\end{table}

\subsection{Effect of point features on scene reconstruction}
\label{subsec:ablation_point_feature_scene}
\captionsetup[subfigure]{position=bottom, labelformat=parens, 
justification=centering}
\begin{figure*}
\centering
\subfloat[]{
  \includegraphics[width=0.37\linewidth,height=0.22\textheight]{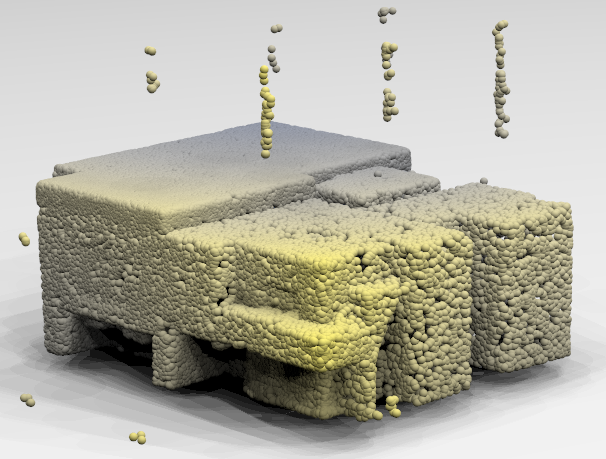} \label{subfig:ablation_ndf_old}} 
\subfloat[]{
  \includegraphics[width=0.37\linewidth,height=0.22\textheight]{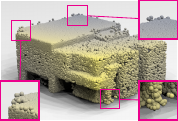} \label{subfig:ablation_pvf_vox}}
  \par \vspace{-1em}
\subfloat[]{
  \includegraphics[width=0.37\linewidth,height=0.22\textheight]{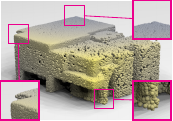} \label{subfig:ablation_pvf}} 
\subfloat[]{
  \includegraphics[width=0.37\linewidth,height=0.22\textheight]{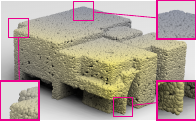} \label{subfig:ablation_gt}}
\caption{An ablation study showing the effectiveness of point features during
training. To reconstruct a scene from the Gibson Environment
\cite{xia2018gibson} dataset, we used (a) the NDF \cite{chibane2020neural}
baseline and (b) \modelname$_\text{wp}$ with our inference algorithm
(Alg.~\ref{alg:inference}). The \modelname\ reconstruction results are shown in
(c) and the ground truth is displayed in (d). Notice that
Alg.~\ref{alg:inference} by itself can reduce the number of outliers. However,
when point features are included during training, our reconstruction results (c)
are closer to the ground truth (d) and achieve more accurate details with far
fewer outliers.}
\label{fig:ablation_point_feature_scene}
\end{figure*}

To test the effectiveness of point features on scene reconstruction, we used
Alg.~\ref{alg:inference} to infer the surface for both \modelname\ and
\modelname$_{\text{wp}}$. The reconstruction results are displayed in
Fig.~\ref{fig:ablation_point_feature_scene}. Compared to the baseline,
Alg.~\ref{alg:inference} by itself can reduce the number of outliers without
point features (Fig.~\ref{subfig:ablation_pvf_vox}). However, when point
features are included during training the reconstruction results
(Fig.~\ref{subfig:ablation_pvf}) are closer to the ground truth
(Fig.~\ref{subfig:ablation_gt}), and therefore more accurate details with less
outliers are realized. 


\begin{figure*}
\centering
\includegraphics[scale=0.29]{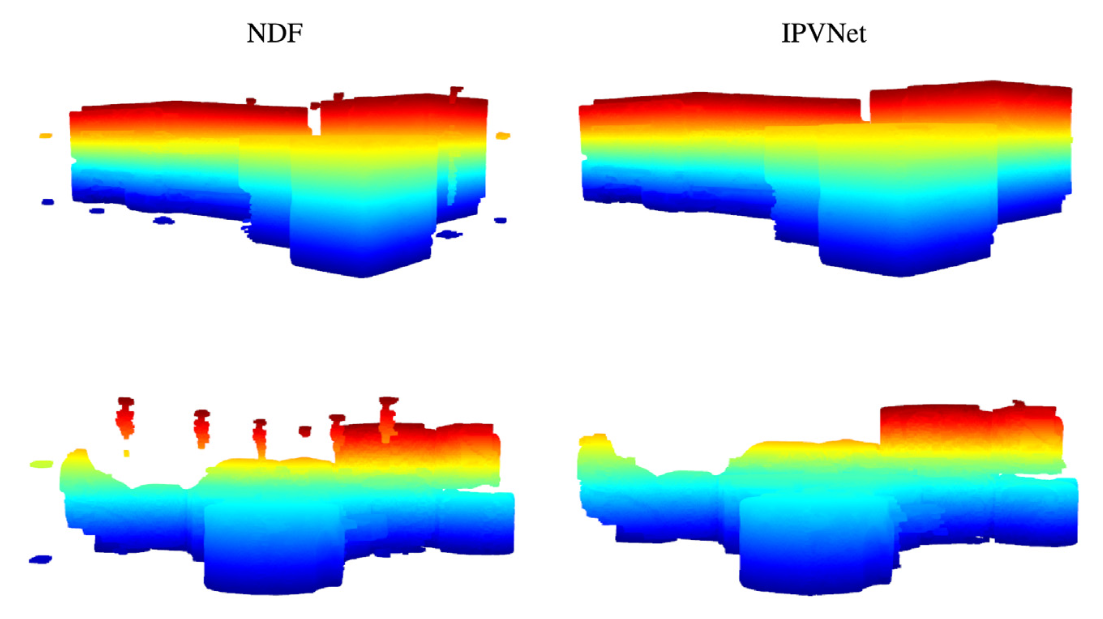}
\caption{Reconstruction results after the NDF \cite{chibane2020neural} baseline
has been filtered using the input coordinate range as the distance threshold,
and \modelname\ without any filtering. NDF still includes outliers due to the
surface curvature whereas the \modelname\ reconstruction consists of
\textit{significantly less} outliers without any filtering.}
\label{fig:post_processing}
\end{figure*}

\subsection{Post-processing outlier removal}
\label{sec:post-processing_outlier_removal}
To provide a comparison of \modelname\ against a naive post-processing step, we
filter the baseline reconstruction using the coordinate range of the input point
cloud as the distance threshold. The qualitative results of this experiment are
recorded in Fig.~\ref{fig:post_processing}. It is critical to note that naive
post-processing cannot remove all the outliers due to their existence near
areas of surface curvature.

\section{Limitations and future directions}
\label{sec:limitations_and_future_directions}
Despite the fact that the UDF function is capable of reconstructing multiple
complex surfaces, the requirement of projecting the query points several times
makes the surface inference time long. Our point-voxel formulation may also be
beneficial for other implicit techniques such as occupancy and signed distance
prediction. We aim to investigate these directions in future work.

\section{Conclusion}
\label{sec:conclusion}
In this paper we introduced \modelname, a novel approach that implicitly learns
from raw point and voxel features to reconstruct complex open surfaces. To
improve the reconstruction quality, we make use of raw point cloud data jointly
with voxels to learn local and global features. Not only have we showed that
\modelname\ outperforms the state of the art on both synthetic and real-world
data, but we also demonstrated the effectiveness of point features on 3D
reconstruction through ablation studies. Furthermore, we developed an inference
module that extracts a zero level set from a UDF and drastically reduces the
amount of outliers in the reconstruction. We believe \modelname\ is an important
step towards reconstructing open surfaces without losing details and introducing
outliers, and we hope that our work will inspire more research in this area.

\section*{Declaration of competing interests}
The authors declare that they have no known competing financial interests or
personal relationships that could have appeared to influence the work reported
in this paper.

\section*{Data availability}
Data will be made available on request.

\section*{Acknowledgments}
The authors acknowledge the Texas Advanced Computing Center (TACC) at The
University of Texas at Austin for providing software, computational, and
storage resources that have contributed to the research results reported within
this paper. 

\bibliographystyle{model1-num-names.bst}
\bibliography{ipvnet_learning_implicit_point-voxel_features_for_open-surface_3d_reconstruction}

\end{document}